\documentclass[sn-mathphys,Numbered]{sn-jnl}% Math and Physical Sciences Reference Style
\usepackage{graphicx}%
\usepackage{multirow}%
\usepackage{amsmath,amssymb,amsfonts}%
\usepackage{amsthm}%
\usepackage{mathrsfs}%
\usepackage[title]{appendix}%
\usepackage{xcolor}%
\usepackage{textcomp}%
\usepackage{manyfoot}%
\usepackage{booktabs}%
\usepackage{algorithm}%
\usepackage{algorithmicx}%
\usepackage{algpseudocode}%
\usepackage{listings}%
\usepackage{url}
\usepackage{hyperref}
\usepackage{color,soul}
\usepackage{comment}
\usepackage{anyfontsize}

\makeatletter
\renewcommand\NAT@open{}  % Rimuove parentesi di apertura
\renewcommand\NAT@close{} % Rimuove parentesi di chiusura
\renewcommand{\cite}[1]{\textsuperscript{\citep{#1}}}
\makeatother
\begin{document}

\title[ ]{\bf Domain knowledge-guided machine learning framework for state of health estimation in Lithium-ion batteries } 
\author[1,2]{\fnm{Andrea} \sur{Lanubile}}\email{lanubile@stanford.edu}
\author[1,2]{\fnm{Pietro} \sur{Bosoni}}\email{pbosoni@stanford.edu}
\author[2]{\fnm{Gabriele} \sur{Pozzato}}\email{pozzato.g@gmail.com}
\author[2]{\fnm{Anirudh} \sur{Allam}}\email{aallam@stanford.edu}
\author[3]{\fnm{Matteo} \sur{Acquarone}}\email{matteo.acquarone@polito.it}
\author*[2]{\fnm{Simona} \sur{Onori}}\email{sonori@stanford.edu}

\affil[1]{These authors contributed equally}
\affil[2]{\orgdiv{Energy Science \& Engineering}, \orgname{Stanford University}, \orgaddress{\street{367 Panama Mall}, \city{Stanford}, \postcode{94305}, \state{CA}, \country{USA}}}
\affil[3]{\orgdiv{Energy Department}, \orgname{Politecnico di Torino}, \orgaddress{\street{24 Corso Duca degli Abruzzi}, \city{Torino}, \postcode{10129}, \country{IT}}}

\abstract{

	Accurate estimation of battery state of health is crucial for effective electric vehicle battery management. Here, we propose five health indicators that can be extracted online from real-world electric vehicle operation and develop a machine learning-based method to estimate the battery state of health. 
	The proposed indicators provide physical insights into the energy and power fade of the battery and enable accurate capacity estimation even with partially missing data. 
	Moreover, they can be computed for portions of the charging profile and real-world driving discharging conditions, facilitating real-time battery degradation estimation. The indicators are computed using experimental data from five cells aged under electric vehicle conditions, and a linear regression model is used to estimate the state of health. The results show that models trained with power autocorrelation and energy-based features achieve capacity estimation with maximum absolute percentage error within 1.5\% to 2.5\% .
}


\maketitle
   
\section*{Introduction}
% Introduction -> greenhouse, regulations, …
%Climate change and global warming are threatening the future of our environment and society, and immediate action is needed to limit and control greenhouse gas emissions.  The transition towards a zero-emission economy requires a readaptation of the transportation sector, which is responsible for 11.9\% of the total global emissions\cite{ritchie2020}. In this framework,  governments and policymakers are implementing regulations to limit vehicular emissions and support the penetration of electrified vehicles\cite{bib4,bib5},  with USA, Europe, and China aiming at net zero vehicular emissions by 2050, 2035, and 2060, respectively\cite{bib2}.

The pressing concern of global warming is driving a global shift towards electrified mobility. 
With the transportation sector contributing to approximately 12\% of all global emissions\cite{bib29}, adjustments are required in order to transition to a zero-emissions energy sector.
Studies by the Intergovernmental Panel on Climate Change\cite{bib29} and the International Energy Agency\cite{bib30} emphasize the critical need for clean transportation solutions to address the urgent issue of climate change.  This has driven governments and policymakers to innovate and collaborate in advancing electric vehicle (EV) technologies.

% SOH 
Lithium-ion batteries (LIBs) are the preferred energy storage technology for EVs  due to their superior power and energy density,  which enables longer driving ranges compared to other battery technologies\cite{LIB-review}. 
For a compelling and sustainable EV mass market,  accurate state of health (SOH) estimation\cite{nuhic2013health} and remaining useful life (RUL)\cite{bib13} prediction of LIB systems are essential.
Existing methods for SOH estimation and RUL prediction can be broadly divided into model-based and data-driven approaches. Model-based estimation approaches rely on empirical or equivalent circuit models (ECMs), or electrochemical models, and formulate estimation algorithms around them.  Various ECM-based filters for SOH estimation have been proposed in the literature,  including Extended Kalman Filter\cite{PletBook2}, dual and joint Extended Kalman Filter\cite{bib39}, Unscented Kalman Filter\cite{UKFECM}, Adaptive Extended Kalman Filter\cite{AEKFTaborelli}, Particle Filter\cite{PFECM} and genetic algorithms\cite{bib35}.

Other methods for SOH estimation and RUL prediction, utilizing empirical degradation models, include Unscented Kalman Filters\cite{miao2013remaining} and Particle Filters\cite{xing2013ensemble}. In a Bayesian Monte Carlo approach\cite{he2011prognostics},  the parameters of an empirical capacity model are updated to compute the posterior probability density function for capacity fade prediction. 
Despite their simplicity, these methods lack explicit physical understanding and require significant calibration effort. Also, electrochemical battery models\cite{bib32,bib33,White_SOH} which demand increased computational power, have been employed, and adaptive observers based on the enhanced single particle model\cite{DiDomenico_SOH,AllamESPM} have been tested in a battery-in-the-loop setup.  

With the advancement in cloud computing technologies and Internet of things, data-driven methods for battery SOH estimation, such as linear regression, gaussian process regression, support vector machine, or artificial neural network, have gained traction in recent years\cite{DataDriven_Recap}.
For instance, multiple linear regression models have been trained using descriptive features of the voltage distribution\cite{vilsen2021battery} or incremental capacity curves\cite{lin2020soh} to predict capacity fade and resistance increase. %The model consistently achieves an absolute percentage error of less than 5\%.
Among the more sophisticated prediction models, Gaussian process regression models have been used for capacity estimation\cite{bib38}, taking as inputs different statistical features extracted from the charging curves\cite{yang2018novel}. %Liu  used a Gaussian Process Functional Regression model to achieve multiple-step-ahead SOH prognostic for LIBs.
Furthermore, neural networks\cite{yang2017neural} are used to establish relationships between input features, such as equivalent circuit model parameters and state of charge (SOC), and battery capacity fade. Regarding RUL prediction, support vector machines\cite{nuhic2013health} and random forests\cite{mansouri2017remaining} are utilized. These methods are effective in forecasting the remaining operational lifespan of batteries based on historical data and operational conditions.
Battery SOH estimation works can be classified into three primary categories based on the dataset used for development.  The first category includes datasets acquired from field operations, which accurately reflect the aging phenomena affecting batteries in real-world EVs driving\cite{allamnc}. However, a challenge with these datasets is the absence of a baseline for evaluating SOH. To address this limitation, several studies\cite{YANG2023107426,doi:10.1177,HONG2021125814}  utilize internal resistance as a direct metric for assessing battery SOH. Alternatively, some studies\cite{bib26} suggest using the peak values derived from incremental capacity curves to overcome this challenge.
However, these metrics can be challenging to evaluate due to their strong dependence on operating conditions, such as temperature.   Capacity fade is often measured using Coulomb counting\cite{bib24,bib25}, which involves integrating battery management system (BMS) current over a limited SOC window. This method, however, may produce inaccurate results due to high sensor noise and quantization in-vehicle sensors.
Conversely, tests are conducted in 
a temperature-controlled environment\cite{bib27} to ensure consistent capacity measurements that serve as ground truth. Here,  SOH is estimated using supervised learning models that directly utilize BMS signals—such as voltage, current, SOC, and pack temperature—as inputs. 
The second category of SOH estimation works relies on datasets collected in laboratory settings\cite{yang2018novel, zhou2016novel, wu2016novel, CHUEHdataset, bib40}. In these datasets, cells are cycled with current profiles that do not accurately represent actual EV battery operation. As a result, features computed using these datasets are not, in general,  transferable nor generalizable to real-world applications.
The third category of datasets utilizes data collected in laboratory settings aimed to mimic EV real-use case scenarios. Examples include ARTEMIS\cite{bib41} and the Urban Dynamometer Driving Schedule (UDDS)\cite{pozzato2022lithium}. These datasets provide more realistic conditions for testing and developing battery SOH estimation algorithms, still providing ground truth capacity through periodic reference performance tests (RPTs). 

Using these datasets, different machine learning algorithms, such as support vector machine\cite{nuhic2013health}, Gaussian process regression, or neural network\cite{bib44}, have been used to estimate SOH for batteries undergoing EV driving cycles using statistical features from current, voltage, and temperature signals as input features.
Other studies have proposed various physics-based health indicators to estimate battery capacity fade, e.g., features derived from the ECM\cite{LYU2022114500} or the time taken for voltage to rise from a low to a high level during the charging process\cite{Shi2022}.Another crucial physical quantity linked to battery aging is internal resistance, with several nuanced indicators proposed in the literature\cite{Chen2018, Lin2023}. Most of these indicators are computed using ECMs and algorithms such as recursive least squares.  However, these methods typically come with increased computational requirements. Another approach\cite{patent2,allamnc} involves evaluating the indicators during the vehicle acceleration (discharge) and braking (charge), requiring less computational power and facilitating its real-time implementation and integration within the vehicle BMS. Another physics-based SOH indicator is charging impedance\cite{allamnc},  which combines variations in electrolyte resistance, charge transfer resistance, and polarization due to aging. This feature can be extracted from the initial portion of the charging phase\cite{Cui2018}.  
Additionally, the energy during charging and discharging can offer valuable insights into battery degradation. Energy metrics are typically calculated over extended portions of full charging profiles to effectively estimate capacity fade\cite{Cai2022,Gong2022,Peng2023}. 

The health features used in previous studies are typically based on idealized constant charging and discharging profiles. However, these profiles do not accurately reflect how electric vehicles are charged and discharged in real-world conditions. Most research has focused on extracting health indicators during complete, repetitive charging cycles, where the battery is charged from a low SOC to a high SOC.  In reality, charging patterns are much more variable, and it's uncommon for batteries to go through full cycles or always follow the same charging profile. This discrepancy makes it difficult to apply findings from controlled experiments directly to real-world EV use.

The contributions of this work are the following.
First, our work systematically formulates various SOH indicators based on domain knowledge and proposes a framework for their integration into BMS. The proposed SOH indicators include: power autocorrelation, resistance, charging impedance, energy during charging, and energy during discharging. Second, unlike previous research\cite{KHALEGHI2019113813} that focused on voltage signal autocorrelation, here, the power autocorrelation is used to quantify the battery's power-delivery capability over time. Additionally, the proposed SOH indicators are derived from an experimental dataset\cite{pozzato2022lithium} that replicates real-world EV battery operation. Unlike most prior studies\cite{Cai2022,Gong2022,Peng2023} that rely on constant current discharging profiles, in this work, energy consumption is evaluated during discharging under realistic EV driving cycles. Moreover,  a windowed approach is proposed to assess energy consumption during charging, thereby improving the effectiveness of the energy as an indicator of battery health, especially in scenarios involving partial charging.  Furthermore, we modify the formulation of the charging impedance indicator\cite{allamnc} by calculating it over an optimized voltage window and averaging the values within this range to improve accuracy and reliability in assessing battery health during partial charging. Through correlation analysis, power autocorrelation, energy during charging, and energy during discharging emerge as the most effective indicators for capacity estimation. 
It is worth noting that the proposed SOH indicators are agnostic to specific battery chemistries. Moreover, they operate independently of cumulative data such as total aging cycles or ampere-hour throughput. This design choice helps mitigate inaccuracies that could arise from sensor errors or insufficient data.  
These indicators can be easily evaluated during EV operation. This makes them suitable for real-time deployment and integration into existing BMS strategies.

The battery capacity is estimated through the machine learning pipeline shown in Supplementary Figure S\ref{fig:ml_pipeline} where  1) SOH indicators are first extracted from the experimental dataset\cite{pozzato2022lithium},  2) the correlation between the indicators and SOH is analyzed through a regression analysis,  and 3) a linear regression model (LRM) is trained to estimate capacity fade. The results show that models trained using power autocorrelation and energy-based features obtain capacity estimation with absolute percentage errors (APE) ranging between 1.5\% and 2.5\%.

Previous works that used linear regression models to estimate battery SOH\cite{vilsen2021battery}\cite{lin2020soh},  are based on features selected over simplistic charge/discharge profile not representative of EV driving.

\section*{Results} \label{sec:results}
The SOH indicators are extracted using data from five Nickel Manganese Cobalt (NMC)/Graphite cells\cite{pozzato2022lithium} (reported in Table \ref{tab:cells} and detailed in Sec. \nameref{sec: dataset}). In this section, a thorough analysis of each indicator and regression analysis is carried out, and the estimation results obtained by the linear regression model are shown.

\renewcommand{\arraystretch}{1.4}

\subsection*{SOH indicators analysis}
\label{sec:indicators_analysis}

\subsubsection*{Power autocorrelation function} \label{sec:PowerAutocorrelation}
The autocorrelation function of the battery's power signal,  % ($P =V \cdot I$), 
evaluated during discharge,  has shown to offer valuable insights into the battery SOH\cite{KHALEGHI2019113813}.
%It's worth noting that, in this study, the discharge phase always follows the UDDS profile, which is consistently repeated until the cell reaches 20\% of SOC. 
Assuming that discharging occurs periodically with an identical current profile, the power autocorrelation $P_{\mathrm{Autocorr}}$  indicator varies as the battery ages, providing a method to monitor battery health.

\begin{figure}
	\centering
	\includegraphics[width = 1\textwidth]{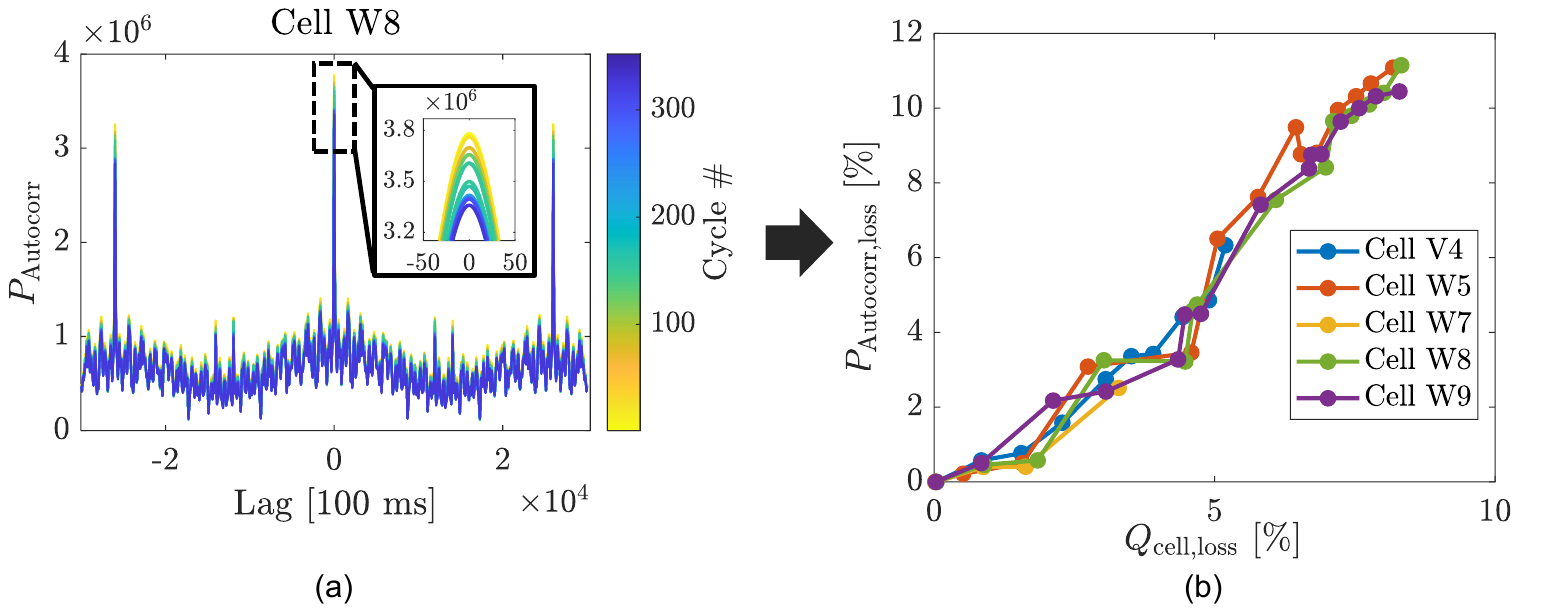}
	\caption{}
 	% \caption{\textbf{Power autocorrelation.} \textbf{a} Power autocorrelation ($P_{\mathrm{Autocorr}}$) profile calculated over discharge profiles of cell W8 throughout  its lifetime. \textbf{b} Percentage decrease of the peak at null delay ($P_{\mathrm{Autocorr,loss}}$) is plotted against the percentage capacity loss ($Q_{\mathrm{cell,loss}}$).}
	\label{fig:PowerAutocorrelation}
\end{figure}

Of particular interest is the change in the central peak of the power autocorrelation function (see Fig. \ref{fig:PowerAutocorrelation}(a)). The reduction in this peak,  defined as $ P_{\mathrm{Autocorr,loss}} = (P_{\mathrm{Autocorr,fresh}} - P_{\mathrm{Autocorr,i}})/P_{\mathrm{Autocorr,fresh}} \cdot 100$) where $P_{\mathrm{Autocorr,fresh}}$ is the central peak value for the fresh cell and $P_{\mathrm{Autocorr,i}}$ is the central peak value during cycle $i$, correlates  with a decrease in capacity. This relationship is illustrated in Fig.\ref{fig:PowerAutocorrelation}(b), where the power autocorrelation loss shows a strong linear relation with capacity loss. Capacity loss is calculated as $ Q_{\mathrm{cell,loss}} = (Q_{\mathrm{cell,fresh}} - Q_{\mathrm{cell,i}})/Q_{\mathrm{cell,fresh}} $, with $Q_{\mathrm{cell,fresh}}$ and $Q_{\mathrm{cell,i}}$ representing the fresh cell capacity and the cell capacity at cycle $i$, respectively.

Despite the promising potential of this indicator for estimating capacity loss, it is important to highlight that the periodicity of the current profile may not hold true in real-driving conditions. Nevertheless, this study suggests that this indicator can be engineered in an offline setting, for example, as part of onboard diagnostics routines. In this work, the power autocorrelation function proves to be an effective SOH indicator given consistent usage of the UDDS discharge profile.

%In this work, we show that $P_{Autocorr}$ is a valuable indicator thanks to the fact that the battery testing current signal (UDDS) is repeated during every discharge phase (by design).

\subsubsection*{Resistance}
Abrupt charge and discharge events, related to braking and acceleration maneuvers, respectively, offer the opportunity to evaluate the battery's internal resistance\cite{allamnc}.
As the battery ages, various factors such as electrode degradation, electrolyte breakdown, and formation of passivation layers contribute to an increase in its internal resistance,  $R$.  This increase limits the flow of ions within the battery, reducing conductivity and affecting the battery's power output capability. As resistance increases, less power can be delivered to the motors due to higher Joule losses.

Demanding acceleration and braking events lead to changes in the battery current, referred to  as current peaks. The resistance is calculated at each discharge current peak corresponding to an acceleration event over the discharging phase of the aging cycle, as described in Sec. \nameref{sec:SOH_indicator_formulation}. It is important to note that the battery's internal resistance is influenced by factors such as SOC (see Supplementary Figure S\ref{fig:ResistanceSOC}),  $C$-rate, and temperature. For accurate aging assessments in real-world scenarios, resistance should be measured under consistent conditions throughout the battery's lifespan. In this work, temperature effects on this indicator are not studied since the cells are maintained in a controlled temperature environment.

\begin{figure}
	\centering
	\includegraphics[width = 1\textwidth]{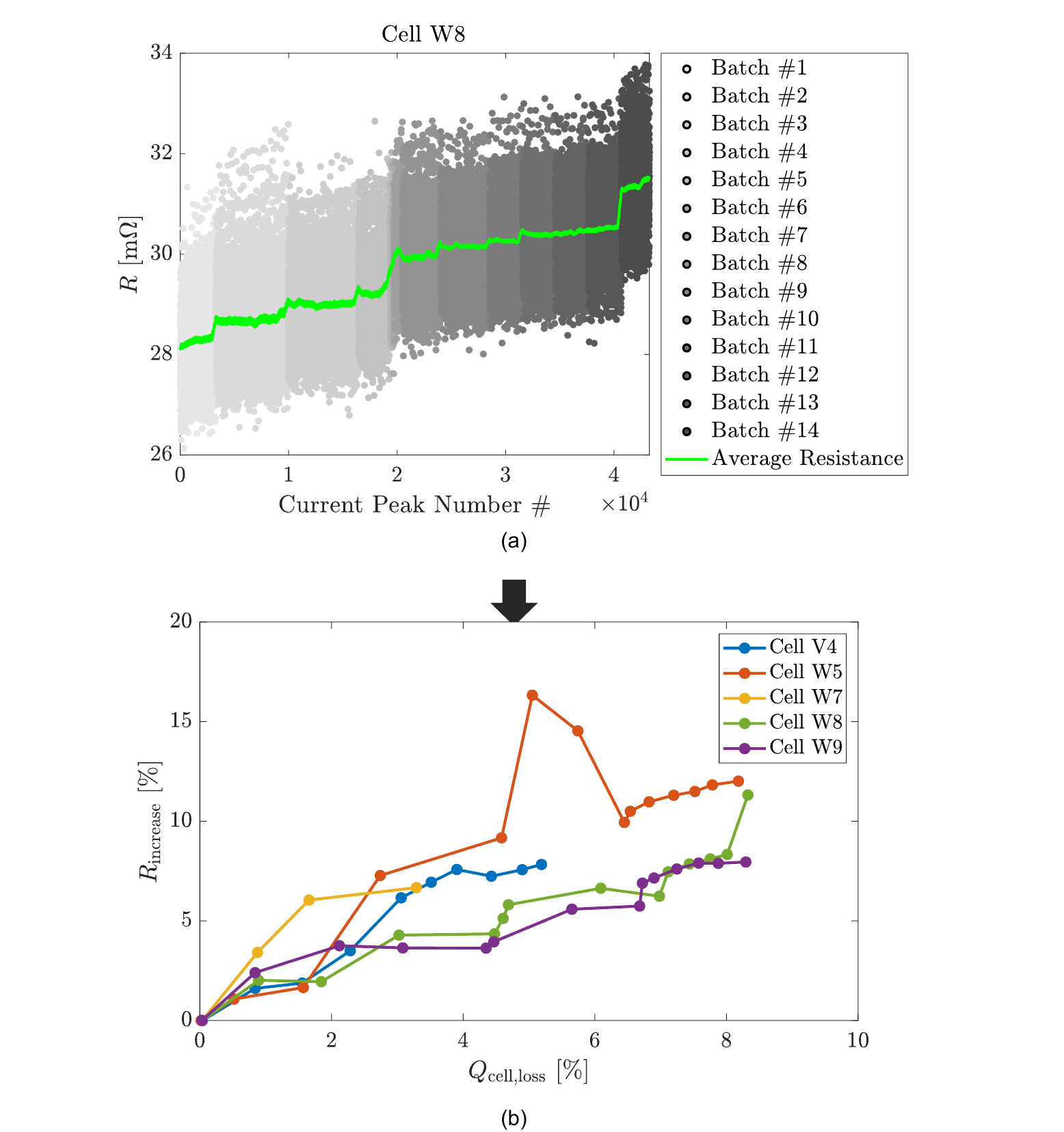}
	\caption{}
	% \caption{ \textbf{Resistance.} \textbf{a} Internal resistance ($R$) is plotted as a function of the current peak number during the discharge phases throughout the cell’s life span for cell W8. Shades of gray represent different batches of aging cycles. Batch $j$ is defined as the period between  the $j$th  and the $(j+1)$th RPT. Green points indicate the average internal resistance. \textbf{b} Percentage increase in internal resistance ($R_{\mathrm{increase}}$) relative to capacity loss ($Q_{\mathrm{cell,loss}}$) for all five cells.}
	\label{fig:Resistance} 
\end{figure}

A single resistance value is computed by averaging the resistances calculated during each discharge phase, which consists of multiple concatenated UDDS cycles between two charging phases, as further detailed in Sec. \nameref{sec: dataset}, to minimize noise in the resistance, as shown in Fig. \ref{fig:Resistance}. This method effectively minimizes noise and variations in resistance measurements, offering a more consistent and representative value to assess battery health. Fig. \ref{fig:Resistance}(a) highlights the importance of determining the average resistance. Despite the large standard deviation observed in the distribution of internal resistances for each discharge event, the average values, represented by green points, clearly exhibits an increasing trend as the battery ages.

Additionally, Fig. \ref{fig:Resistance}(b) shows the percentage increase in average internal resistance for all five cells, correlated with their corresponding capacity losses. This increase is calculated as $R_{\mathrm{increase}} = (R_{i} - R_{\mathrm{fresh}}) / R_{\mathrm{fresh}} \cdot 100$, where $R_{\mathrm{fresh}}$ represents the average internal resistance measured during the first discharging phase of the cell, and $R_{i}$ denotes the average internal resistance determined during the discharging at cycle $i$.

\subsubsection*{Charging impedance}
\label{subsub:ChargingImpedance}
\begin{figure}
	\centering
	\includegraphics[width = 1\textwidth]{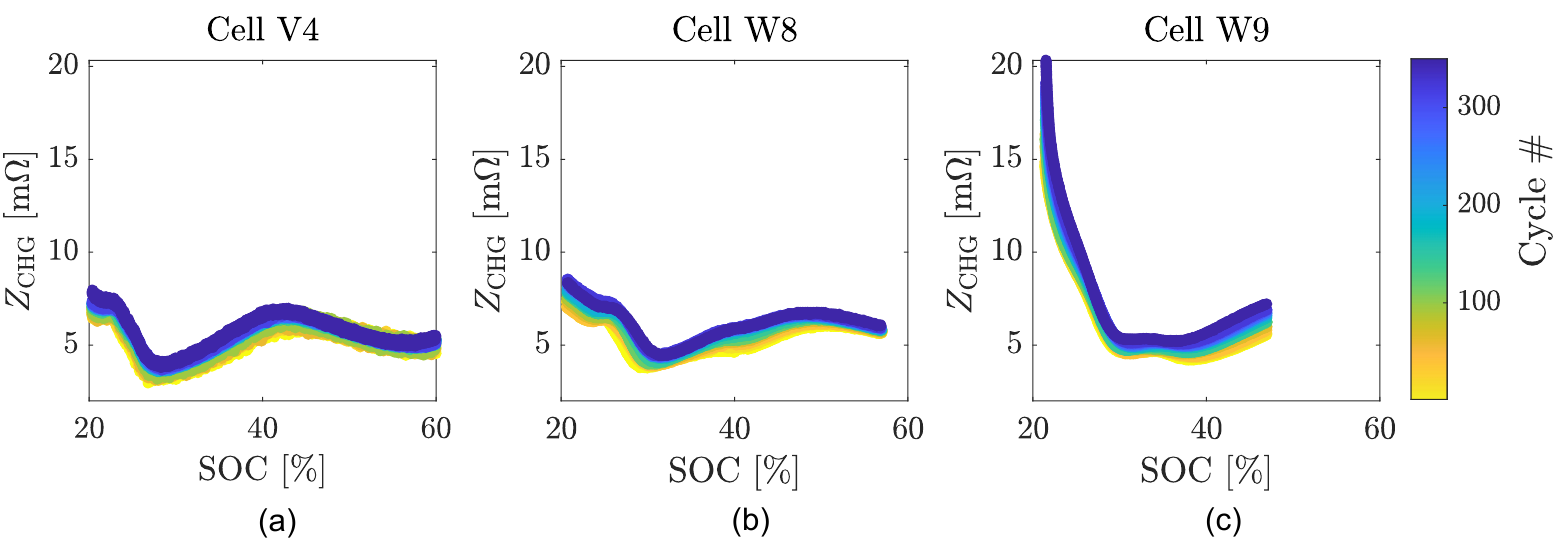}
	\caption{}
	% \caption{\textbf{Charging Impedance ($\mathbf{Z_{\mathrm{CHG}}}$) as function of SOC and cycle number for cells V4 (a), W8 (b), and W9 (c).} The yellow curve represents fresh cell conditions, while the dark blue curve denotes aged cell conditions.}
	\label{fig:pR_withandwithoutOCV} 
\end{figure}

\begin{figure}
	\centering
	\includegraphics[width = 1\textwidth]{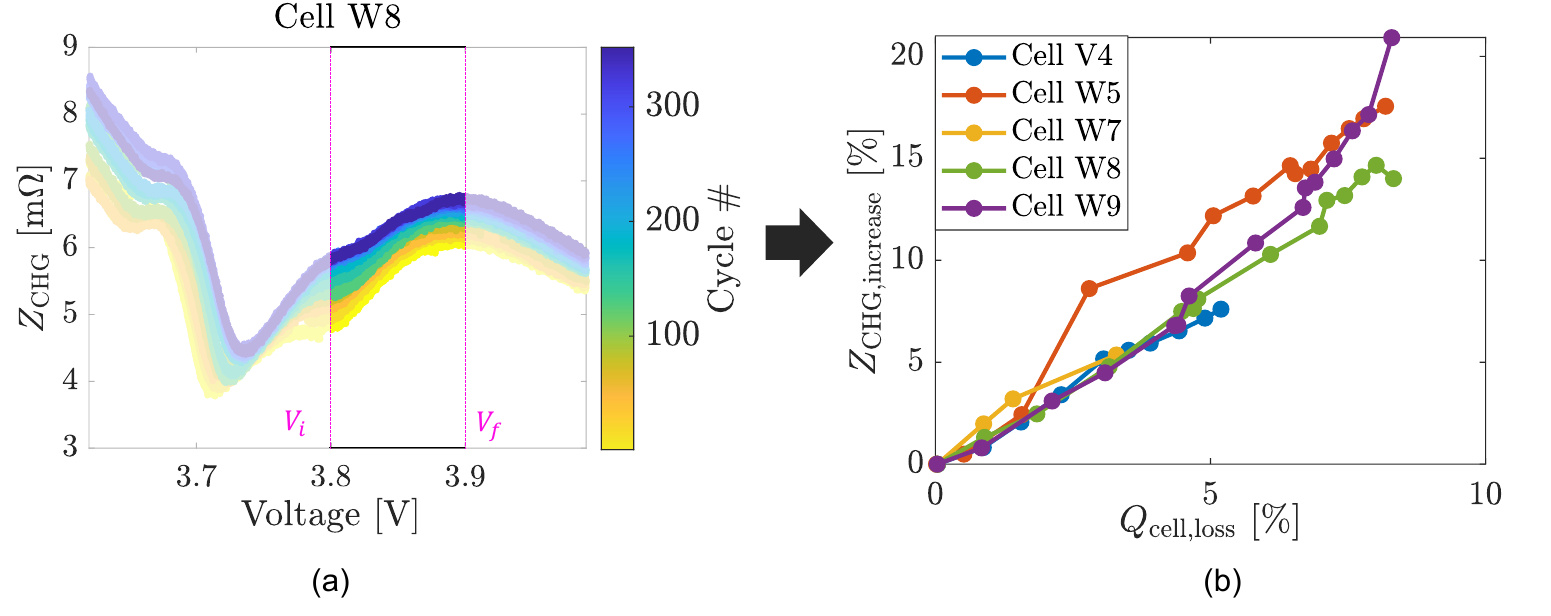}
	\caption{}
	% \caption{\textbf{Charging Impedance.} \textbf{a} Charging Impedance ($Z_{\mathrm{CHG}}$) as a function of voltage for cell W8. The voltage range [$V_{\mathrm{in}}$ = 3.8 V, $V_{\mathrm{fin}}$ = 3.9 V], over which the charging impedance is averaged, is highlighted. \textbf{b} Percentage variation of average $Z_{\mathrm{CHG}}$ ($Z_{\mathrm{CHG,increase}}$) as a function of capacity fade ($Q_{\mathrm{cell,loss}}$).}
	\label{fig:pR_withandwithoutOCV_corr} 
\end{figure}

The charging impedance\cite{allamnc} $Z_{\mathrm{CHG}}$ represents the battery's resistance to the flow of electrons during charging. Variations in  $Z_{\mathrm{CHG}}$ reflect how this resistance evolves as the battery ages. The $Z_{\mathrm{CHG}}$  profiles for three cells (V4, W8, and W9), charged at different $C$-rates, are illustrated in Fig. \ref{fig:pR_withandwithoutOCV} as a function of cell degradation and SOC.
The rising trend of $Z_{\mathrm{CHG}}$ over cells' lifetime aligns with the understanding that, as the battery ages, its overpotential increases due to factors such as the growth of the Solid-Electrolyte Interface, increased of contact resistance, and changes in reaction kinetics and transport dynamics\cite{PletBook2,ovejas2019effects}.   Additionally, it is important to note that the  $Z_{\mathrm{CHG}}$ profiles reach different SOC values at the end of charge (at 4 V). This phenomenon can be attributed to the varying polarization losses resulting from the different $C$-rates used during charging for cells V4, W8, and W9\cite{fly2020rate}. The charging impedance indicator is computed by averaging the impedance within the specific voltage range [$V_{\mathrm{in}}$ = 3.8 V, $V_{\mathrm{fin}}$ = 3.9 V], which is selected through the analysis reported in the Supplementary Notes \nameref{sup:impedance_with_OCV}, with additional details in the Supplementary Notes  \nameref{sup:pseudo-ocv}, \nameref{sup:t-ocv}.
As shown in Fig. \ref{fig:pR_withandwithoutOCV_corr}(b), the increase in charging impedance  ($Z_{\mathrm{CHG,increase}}=(Z_{\mathrm{CHG,i}}-Z_{\mathrm{CHG,fresh}})/Z_{\mathrm{CHG,fresh}}\cdot 100$) is highly correlated with capacity loss across all the battery cells. 
    Therefore, the charging impedance $Z_{\mathrm{CHG}}$ can be used directly as a feature to correlate with capacity loss.

\subsubsection*{Energy during charging}

\begin{figure}
\centering
\includegraphics[width = 1\textwidth]{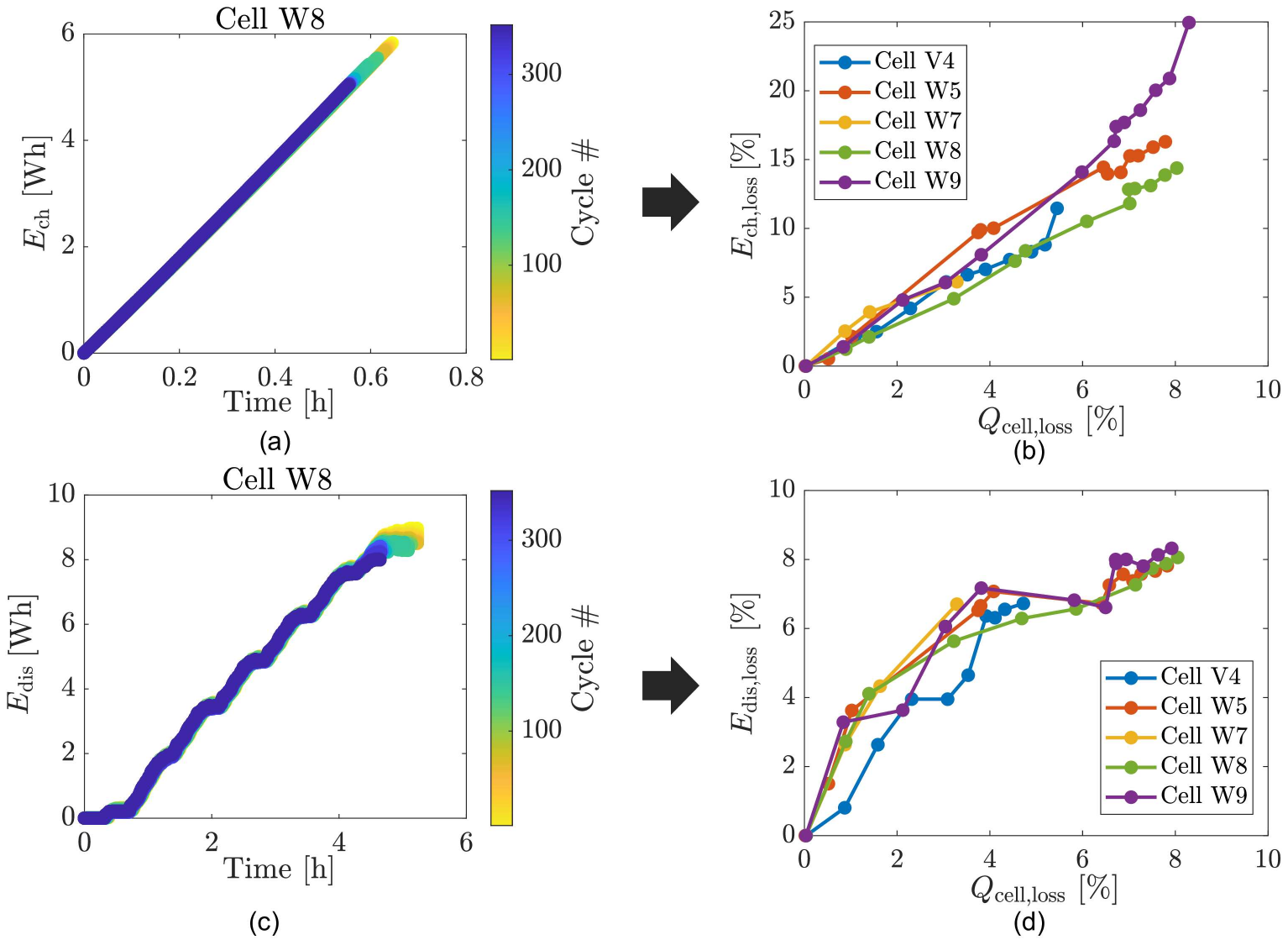}
\caption{}
% \caption{\textbf{Energy during charging and discharging.} \textbf{a}  Energy during charging ($E_{\mathrm{ch}}$) as a function of charging time within the voltage range [$V_{\mathrm{in,ch}}$ = 3.6 V,$V_{\mathrm{fin,ch}}$ = 3.9 V] for cell W8 throughout its cycle life. \textbf{b}  Energy loss during charging ($E_{\mathrm{ch,loss}}$) across all 5 cells shows a linear correlation with capacity loss ($Q_{\mathrm{cell,loss}}$). \textbf{c} Energy during discharging ($E_{\mathrm{dis}}$) as a function of discharging time within the voltage range [$V_{\mathrm{in,dis}}$ = 3.85 V, $V_{\mathrm{fin,dis}}$ = 3.4 V] for cell W8 throughout its cycle life. \textbf{d}  Energy loss during discharging ($E_{\mathrm{dis,loss}}$) for all five cells demonstrates a linear correlation with capacity loss ($Q_{\mathrm{cell,loss}}$).}
\label{fig:energy} 
\end{figure}

The energy during charging indicator, $E_{\mathrm{ch}}$, quantifies the energy stored in the battery during charging.  This is computed by integrating the battery power within a specific voltage range [$V_{\mathrm{in,ch}}$, $V_{\mathrm{fin,ch}}$] (as detailed in Sec. \nameref{sec:SOH_indicator_formulation}). 
Fig. \ref{fig:energy}(a) illustrates $E_{\mathrm{ch}}$ in relation to the charging duration required to reach $V_{\mathrm{fin,ch}}$ from $V_{\mathrm{in,ch}}$.  Fig. \ref{fig:energy}(b) shows the energy during charging over the voltage range [$V_{\mathrm{in,ch}}$ = 3.6 V,$V_{\mathrm{fin,ch}}$ = 3.9 V] as a function of capacity loss. 
The y-axis of Fig. \ref{fig:energy}(b) quantifies the percentage energy loss during charging  for each cell. Energy loss during charging  for each cell is computed as $ E_{\mathrm{ch,loss}} = (E_{\mathrm{ch,fresh}} - E_{\mathrm{ch,}i})/E_{\mathrm{ch,fresh}} \cdot 100 $, where $E_{\mathrm{ch,fresh}}$ is the energy for the fresh cell and $E_{\mathrm{ch,}i}$ is the amount of energy the battery is charged at during aging cycle $i$ of the same cell. These results show that capacity loss is linearly correlated with energy loss during charging over the selected voltage range. %Hence,  this indicator can be directly used as a feature for the estimation of capacity loss.

\subsubsection*{Energy during discharging}

The energy during discharging indicator, $E_{\mathrm{dis}}$,quantifies the energy delivered by the battery during its discharge phase. This is computed by integrating the battery power over a specific voltage range [$V_{\mathrm{in,dis}}$, $V_{\mathrm{fin,dis}}$], as detailed in Sec. \nameref{sec:SOH_indicator_formulation}. 
Fig. \ref{fig:energy}(c) illustrates $E_{\mathrm{dis}}$ in relation to the discharging duration needed to reach $V_{\mathrm{fin,dis}}$ from $V_{\mathrm{in,dis}}$.  Fig. \ref{fig:energy}(d) displays the energy during discharging over the voltage range [$V_{\mathrm{in,dis}}$ = 3.85 V,$V_{\mathrm{fin,dis}}$ = 3.4 V] as a function of capacity loss. 
The y-axis of Fig. \ref{fig:energy}(d) quantifies the percentage energy loss during discharging for each cell. Energy loss during discharging for each cell is computed using $ E_{\mathrm{dis,loss}} = (E_{\mathrm{dis,fresh}} - E_{\mathrm{dis,i}})/E_{\mathrm{dis,fresh}} \cdot 100 $, where $E_{\mathrm{dis,fresh}}$ represents the energy of a fresh cell, and $E_{\mathrm{dis,i}}$ is the energy charged during aging cycle $i$ of the same cell. The results indicate a linear relationship between capacity loss and energy loss during discharging within the selected voltage range.
In real EV scenarios, the variability in discharging rates complicates the consistent computation and monitoring of $E_{\mathrm{dis}}$.  A practical approach is to compare $E_{\mathrm{dis}}$ across driving scenarios with similar driving styles to account for this variability.

\subsubsection*{SOH indicators regression analysis} \label{sec:regression}

\begin{figure}
	\centering
	  \includegraphics[width = 1\textwidth]{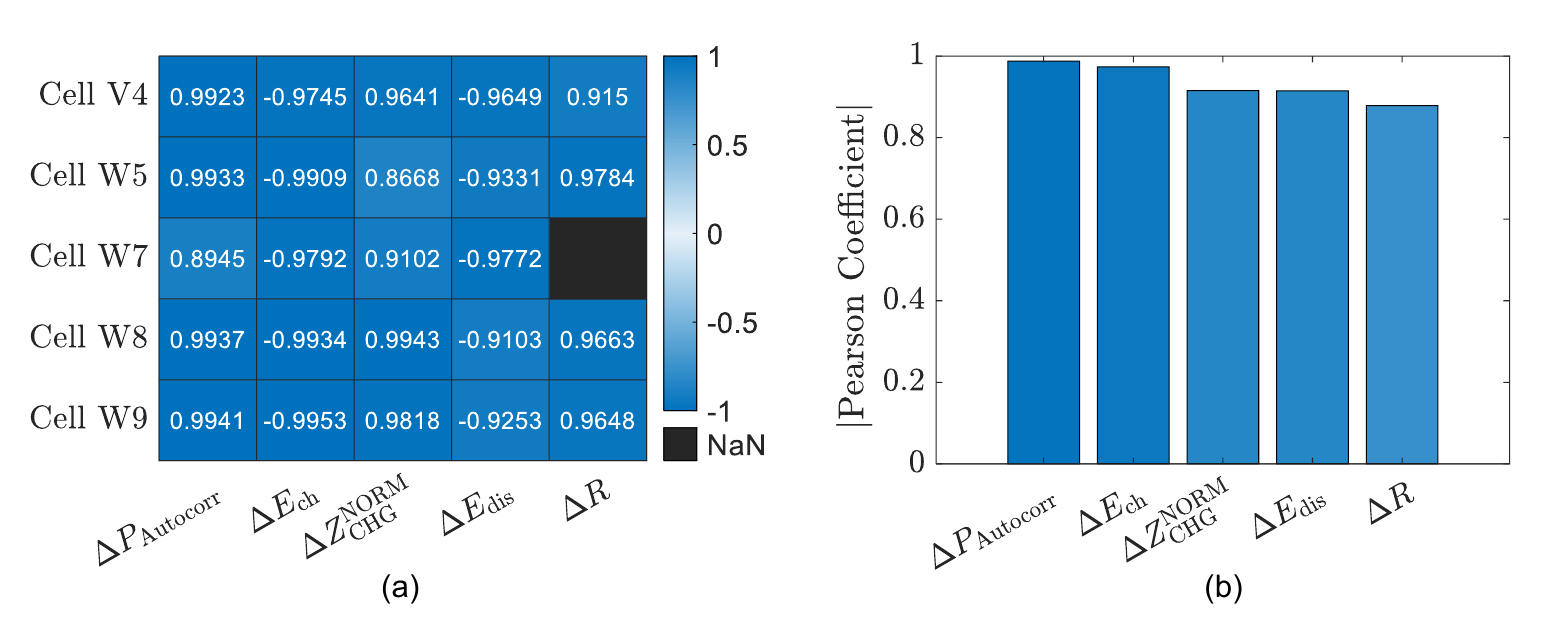}
	  \hfill
	\caption{}
	% \caption{\rev{\textbf{Correlation Analysis.} \textbf{a} Heatmap showing the Pearson's correlation coefficients between incremental power autocorrelation, incremental energy during charging, incremental energy during discharging, normalized incremental charging impedance, and incremental resistance ($\Delta \textit{P}_{\mathrm{Autocorr}}$, $\Delta \textit{E}_{\mathrm{ch}}$, $\Delta \textit{E}_{\mathrm{dis}}$, $\Delta \textit{Z}_{\mathrm{CHG}}^{\mathrm{NORM}}$ and $\Delta R$, respectively) with capacity loss for each individual cell.}  \textbf{b} Histogram illustrating the Pearson's correlation   coefficients between incremental features and capacity loss across all cell data. Note that the correlation between the capacity of cell W7 and $\Delta R$ is not reported due to some computed resistances being deemed unreliable because of data acquisition issues  (Sec. \nameref{sec: dataset} and Supplementary Note \nameref{sup:IncreasedVoltage}), and thus interpreted as outliers during the pre-processing phase.}
	\label{fig:Regression}
\end{figure}

The health indicators are pre-processed according to the pipeline outlined in Supplementary Figure S\ref{fig:ml_pipeline}. This process involves calculating incremental values for each feature: 
$\Delta P_{\mathrm{Autocorr}}$ (power autocorrelation), $\Delta R_{\mathrm{ch}}$ (resistance), $\Delta Z_{\mathrm{CHG}}^{\mathrm{NORM}}$ (normalized charging impedance), $\Delta E_{\mathrm{ch}}$ (energy during charging), and $\Delta E_{\mathrm{dis}}$ (energy during discharging). 
These incremental values are derived by subtracting the initial feature value, measured during the first aging cycle, from the value at each subsequent aging cycle $i$  throughout the cell's life cycle. Additional details are provided in  Sec.~\nameref{sec:methods}.
In this work, we use features' incremental values to simplify the detection of aging trends. For each cell, we assess the correlation between its capacity loss and feature variations   using Pearson's correlation coefficient $r$, defined as: 
\begin{equation}
    r = \frac{\sum_{i=1}^N (X_i - \overline{X})(Y_i - \overline{Y})}{\sqrt{\sum_{i=1}^N (X_i - \overline{X})^2 \sum_{i=1}^N (Y_i - \overline{Y})^2}},
\end{equation}
where $X_i$ represents the value of a specific incremental feature for a given cell at the  $i$-th aging cycle, $Y_i$ is the corresponding capacity loss value for the same cell at that cycle, $\overline{X}$ is the mean of the incremental feature values, $\overline{Y}$ is the mean of the capacity loss values across all cycles, and $N$ is the total number of data points (aging cycles analyzed). The results are shown in the heatmap of Fig. \ref{fig:Regression}(a). Each cell shows a high Pearson's correlation coefficient between capacity loss and each feature, underlying that the variations in these features are consistent indicators of aging across all the cells. 
However, since feature trends can vary across different cells, an additional analysis was conducted to identify features with more generalizable trends. We performed a correlation analysis between the extracted incremental features and capacity fade across all cells. This approach helps identify features that consistently reflect cell aging, regardless of individual cell differences. Fig. \ref{fig:Regression}(b) shows that some indicators generalize better across different cells.

We select features to train an estimation model according to two different cases. In the first case, Power autocorrelation ($P_{\mathrm{Autocorr}}$) is selected, as the sole feature, due to its superior overall performance. In the second case, we choose the best-performing feature for charging (energy during charging, $E_{\mathrm{ch}}$) and the best-performing feature for discharging (energy during discharging, $E_{\mathrm{dis}}$), excluding power autocorrelation.

The strong correlation between the extracted features and capacity fade can be attributed to the physical phenomena driving battery degradation. The linear relationship observed between charging impedance, resistance, and energy features with respect to charge throughput aligns with the linear trend of the capacity fade curve\cite{Yang2017}. 
Given that the cells are cycled within a linear SOC window of 80\% to 20\% at ambient temperature, Solid-Electrolyte Interface layer growth is considered the dominant aging mechanism, leading to a linear capacity decrease trajectory. However, to thoroughly assess the aging modes present in the cells, a post-mortem analysis would be necessary.

%The feature selection for the training of the ML model is further detailed in the following subsection \ref{sec:capacity_estimation_results}. \\
% It should be noted that the resistances computed for cell W7 were disregarded, due to the acquisition problems previously mentioned, which strongly affected the computation of this indicator.
 
\subsection*{SOH estimation}\label{sec:capacity_estimation_results}

In this paper, we use capacity calculated at $C$/20 during RPTs as SOH metric. Additionally, for the purpose of  training the machine learning models, the experimental $C$/20 capacity points are augmented using a linear data augmentation method as discussed in Sec. \nameref{sec:data_augmentation}.

The features selected through the regression analysis are utilized to estimate capacity loss using a data-driven model. The performance of various models, namely, LRM, feed-forward neural networks, autoregressive moving average with extra input, and recurrent neural networks, is compared using the same training and testing datasets, as detailed in Supplementary Note \nameref{sup:model_comparison}. 
Despite its simplicity, the LRM achieves estimation performance comparable to that of more complex models, owing to the strong linear correlation between the SOH indicators and capacity degradation. Therefore, the LRM is chosen for capacity loss estimation due to its lower computational time. Additionally, the LRM has the advantage of requiring fewer parameters to tune and fewer training samples compared to neural network-based models\cite{jiao2020does}.
The LRM is trained using distinct sets of incremental SOH features: first with power autocorrelation, and then with energy during charging and energy during discharging (see Sec. \nameref{sec:regression}). Additionally, the estimation capabilities of the selected features are evaluated in two Scenarios. In Scenario 1 the LRM is trained exclusively on the data from  cell W8 and tested on the other cells. In Scenario 2 the LRM is trained using data from all cells except the test cell. In the second Scenario, for cross-validation, the data is split into two subsets: one for the target cell and another for the remaining cells. The model is trained on the data from the remaining cells and tested on the data from the target cell.

\begin{figure}
	\centering
	  \includegraphics[width = 1\textwidth]{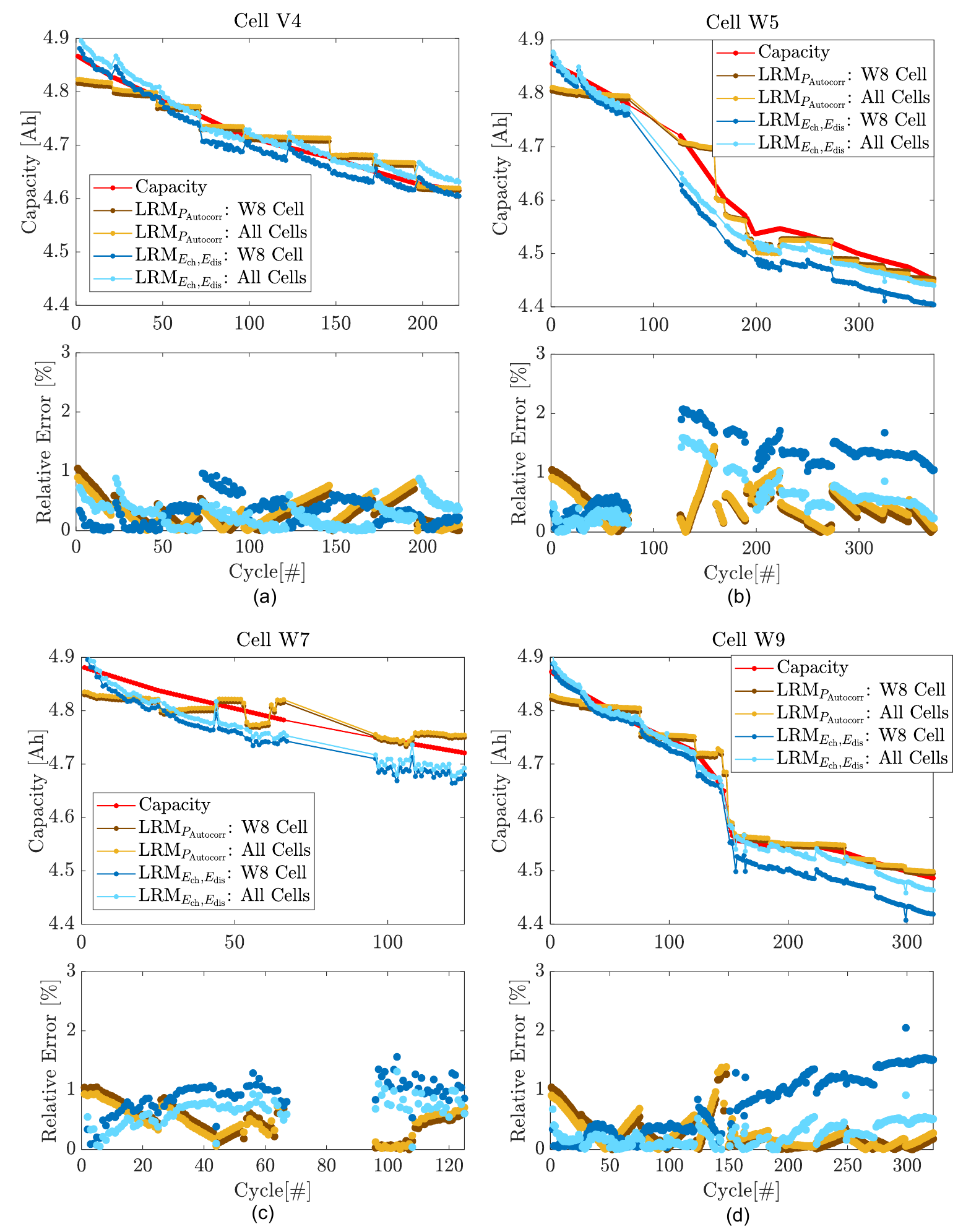}
	\caption{}
	% \caption{\rev{\textbf{SOH estimation results from the LRM using power autocorrelation ($\mathrm{LRM}_{P_\mathrm{Autocorr}}$), and energy during charging and discharging ($\mathrm{LRM}_{E_\mathrm{ch},E_\mathrm{dis}}$) as input features.}} Profiles of capacity loss and estimation error for cells  V4 (\textbf{a}),  W5 (\textbf{b}),  W7 (\textbf{c}), and W9 (\textbf{d}). Augmented Capacity points (obtained as discussed in Sec. \nameref{sec:data_augmentation}) are shown in red. SOH estimation using power autocorrelation as input is shown in brown (with training data from cell W8) and yellow (with training data from all cells except the test cell). The dark blue and light blue lines show SOH estimation using energy features as input, with training data from cell W8 (dark blue) and from all available cells except the test cell (light blue). Gaps in the capacity curves for cells W5 (\textbf{b}) and W7 (\textbf{c}) are due to an increased output voltage profile affecting the reliability of feature values (see Sec. \nameref{sec: dataset} and Supplementary Note \nameref{sup:IncreasedVoltage}). The capacity drop for cell W8 (\textbf{d}) results from issues with the aging protocol implementation.}
	\label{fig:ResultsAutocorrvsEnergies}
\end{figure}

Since the autocorrelation function of the power signal $\Delta \textit{\textbf{P}}_{\mathrm{Autocorr}}$ exhibits the highest correlation with capacity fade, the data-driven model is initially trained using using $\Delta \textit{\textbf{P}}_{\mathrm{Autocorr}}$ as input. Fig. \ref{fig:ResultsAutocorrvsEnergies} displays the capacity estimation results for both  $\Delta \textit{\textbf{P}}_{\mathrm{Autocorr}}$ and the energy-based features. 
In Scenario 1, the training dataset consists solely of data from cell W8, while in Scenario 2, it includes data from all cells except the test cell. The absolute percentage error (as defined in Sec.~\nameref{sec:methods}) remains consistently below 1.5\%, underscoring the relevant information provided by this individual feature.
Moreover, using a more extensive set of training data from multiple cells (Scenario 2) does not  improve estimation accuracy, leading to conclude that  $\Delta \textit{\textbf{P}}_{\mathrm{Autocorr}}$  is effective even with limited data.  However, this feature has limitations in real-world scenarios and is better suited for offline diagnostics rather than online applications. It is also important to note that gaps in the observed capacity curves are due to voltage measurements anomalies, which resulted in unreliable feature values. This irregularity is attributed to unidentified equipment issues, as discussed in Sec.  \nameref{sec: dataset} and detailed further in the Supplementary Note \nameref{sup:IncreasedVoltage}.

\begin{figure}[H]
	\centering
 	\includegraphics[width = 1\textwidth]{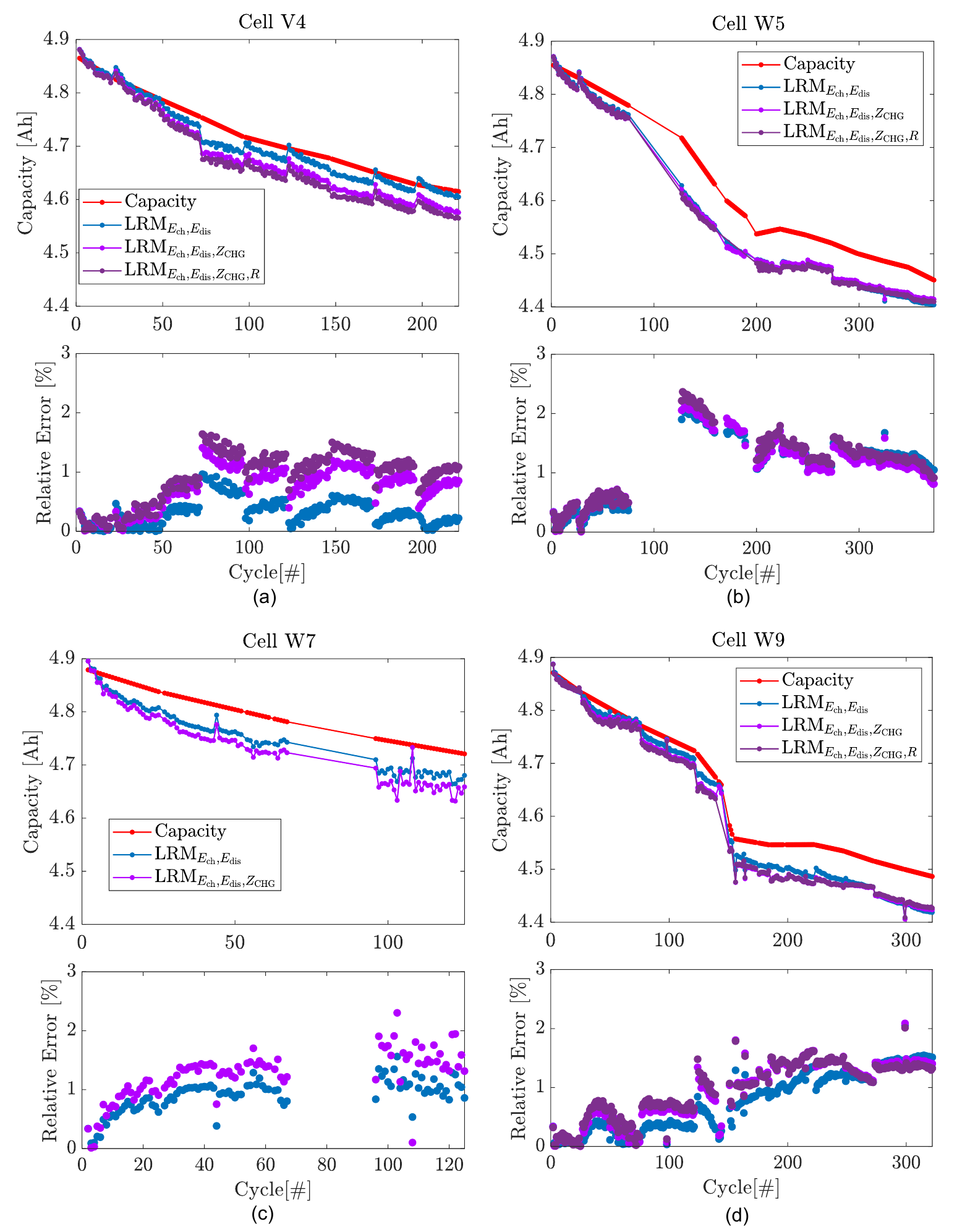}
	\caption{}
	% \caption{\textbf{SOH estimation results from the LRM using charge and discharge energies, charging impedance, and resistance as features.} The capacity loss and estimation error profiles for cells V4 (\textbf{a}), W5 (\textbf{b}), W7 (\textbf{c}), and W9 (\textbf{d}) are shown. Augmented Capacity points (obtained as discussed in Sec. \nameref{sec:data_augmentation}) are shown in red. \rev{Three scenarios are displayed: blue represents the LRM output trained solely with energy during charging and energy during discharging ($\mathrm{LRM}_{E_\mathrm{ch},E_\mathrm{dis}}$); the light purple represents the LRM output trained with energy during charging, energy during discharging along with charging impedance ($\mathrm{LRM}_{E_\mathrm{ch},E_\mathrm{dis},Z_\mathrm{CHG}}$); the dark purple represents the LRM output trained with energy during charging, energy during discharging, charging impedance, and resistance ($\mathrm{LRM}_{E_\mathrm{ch},E_\mathrm{dis},Z_\mathrm{CHG},R}$). All models are trained exclusively using data from cell W8.}}
	\label{fig:Energies_vs_EnergiesImpedanceResistance}
\end{figure}

The LRM is subsequently trained using features that can be calculated during vehicle operation, specifically during driving and charging. 
The features selected for their high linear correlation with capacity during charging and discharging are energy during charging ($\Delta \textit{\textbf{E}}_{\mathrm{ch}}$) and energy during discharging ($\Delta \textit{\textbf{E}}_{\mathrm{dis}}$), respectively.
As illustrated in Fig.~\ref{fig:ResultsAutocorrvsEnergies}, accurate capacity fade estimation is achieved with these features.
Notably, when the LRM is trained using data from only cell W8 (Scenario 1), it achieves an absolute percentage error below $2.5\%$ when tested on data from the other four cells. This result highlights the strong estimation capability of these features even with a limited dataset. For a more comprehensive analysis, the same estimation model is trained using data from multiple cells, leading to improved performance with the larger dataset.
When using data from four cells for training (Scenario 2) and testing on the remaining cell, the absolute percentage error is below $1.6\%$. Notably, the estimation models perform well even for cells like W7 and W5, where some data is missing. This adaptability of the features and estimation models to partially available data is particularly advantageous in real-world scenarios, where acquiring complete EV battery data may not always be feasible.
Moreover, to evaluate if adding extra features alongside the energy-based indicators could enhance model estimation capabilities, the LRM was also trained with incremental resistance and charging impedance included as additional inputs. However, the performance of the model with these additional features was worse than when using only energy during charging and discharging, as shown in Fig. \ref{fig:Energies_vs_EnergiesImpedanceResistance}. This indicates that the inclusion of resistance and charging impedance may introduce more noise than valuable information.
It should be noted that for cell W7, only charging impedance is used as an additional feature, as the resistance data was compromised due to acquisition issues discussed in Sec. \nameref{sec:methods}.
The superior performance of energy during charging and discharging as SOH indicators, compared to the increase in resistance or charging impedance, can be attributed to several factors. Energy loss reflects not only resistance increases but also other factors such as heat generation, electrode degradation, and Solid-Electrolyte Interface formation, which impact overall energy efficiency. Additionally, the integration of the power signal offers a comprehensive measure of battery energy dynamics throughout an entire cycle, whereas resistance and charging impedance are computed over shorter time periods, making them more sensitive to short-term fluctuations.

\section*{Conclusions}
This work extracts and evaluates five knowledge based SOH indicators, demonstrating their effectiveness as inputs to ML models for estimating capacity fade.
The formulation of these indicators is guided by battery domain knowledge, allowing for the quantification of internal state variability due to battery degradation. Since none of the indicators rely on cumulative information (such as cycle number or Ah-throughput), they are suitable for real-world applications even with partial battery history. The high correlation between the indicators and capacity indicates that battery aging mechanisms leading to capacity fade are directly related to energy decrease and impedance rise.
Two subsets of the engineered indicators, i.e.,power autocorrelation, energy during charging, and energy during discharging, were utilized to train the estimation model for accurate cell capacity estimation. Due to their high correlation with capacity fade, combining energy during charging and energy during discharging as inputs results in accurate SOH estimation, with an absolute percentage error consistently below 2.5\%. Conversely, power autocorrelation is the most informative feature, enabling precise capacity fade estimation with an absolute percentage error below 1.5\%, even with limited training data. However, its effectiveness is influenced by the periodicity of discharging events. Consequently, power autocorrelation cannot be directly used as an SOH indicator in real-world driving scenarios but could be incorporated into a diagnostic tool by applying a periodic current signal to the battery when it is not in use.
These findings suggest that domain knowledge-based features have the potential to be used as online tools for real-time capacity estimation.
However, the model's effectiveness may be limited in practical applications. The dataset used in this study does not account for temperature variations or practical discharge events typical in real-world battery usage. Additionally, the current and voltage signals used to extract features have a high signal-to-noise ratio, which may not always be present in EV batteries.
Having demonstrated the potential of these features on the studied dataset\cite{pozzato2022lithium}, further investigations will be conducted using field data as future work. While this study primarily focuses on capacity estimation, utilizing a larger dataset could allow for the application of these indicators in RUL prediction. Extending the method proposed in this paper, these indicators could be integrated into forecasting models, enabling the BMS to anticipate and effectively manage battery capacity degradation.

\section*{Methods} \label{sec:methods}
\subsection*{Cell cycling and experimental dataset}\label{sec: dataset}
%This paper focuses on the computation of SOH indicators incorporating information on battery health.  

The experimental dataset\cite{pozzato2022lithium} used in this work involves INR21700-M50T battery cells with graphite/silicon anode and nickel manganese cobalt oxides (NMC) cathode tested over a period of 30 months. 
For each cell, periodic RPTs, including  \textit{C}/20 capacity tests, Hybrid Pulse Power Characterization,  and Electrochemical Impedance Spectroscopy, were conducted to assess the battery aging from fresh conditions.
The cells underwent aging cycles as described in Supplementary Figure S\ref{fig:sel_cycle}. Each cycle includes a Constant Current-Constant Voltage (CC-CV) charge phase followed by a discharge phase. 
Specifically, there are two charge phases. Once the batteries reach 20\% SOC (from the discharge phase), they are charged through the CC-a phase (at different $C$-rates) until reaching 4 V. They then continue charging at \textit{C}/4 until 4.2 V, followed by the CV phase until the current drops below 50mA.
The discharge phase, using concatenated UDDS driving profiles, simulates EV battery discharging, reducing the cell's SOC from 80\% to 20\%. Aging cycles conducted between the $j^{\mathrm{th}}$ and $(j+1)^{\mathrm{th}}$ RPTs for each cell are grouped into the $j^{\mathrm{th}}$ batch of aging cycles. Supplementary Table S\ref{suptab1} details the number of aging cycles in each batch for all cells used in this study.
Among the ten cells (G1, V4, V5, W3, W4, W5, W7, W8, W9, W10) in the dataset, five (V4, W5, W7, W8, W9) are used in this study, as detailed in Table \ref{tab:cells}. 
The remaining cells were excluded for the following reasons. Cells W3, W10, and G1 were charged using a fast-charging 3\textit{C} current profile during the CC-a phase, resulting in a very short charging duration interval that hindered feature extraction. Cell V5  was excluded due to insufficient aging, having undergone only 59 cycles with less than a 3\% capacity decrease from the beginning of life. Cells W4, W5, and W7 were reported to have voltage measurements anomalies due to experimental issues, as noted in the “README” file of the dataset\cite{pozzato2022lithium} and detailed in Supplementary Note \nameref{sup:IncreasedVoltage}. Specifically, cell W4 was affected for 310 cycles out of the total 760.

\subsection*{Data augmentation approach}\label{sec:data_augmentation}
This work uses capacity to describe battery SOH. Given the limited number of RPTs, we have adopted an approach that uses  data augmentation with linear interpolation for training purposes. For each cell, to assign a capacity value at every aging cycle $i$ contained in batch $j$, we use the capacity values measured at the $j$-th and $(j+1)$-th RPTs and estimate the capacity for cycle $i$, $Q_i$ as follows:
\begin{equation}
	Q_i = \frac{i - \text{cycle}^{\mathrm{RPT}}_j}{\text{cycle}^{\mathrm{RPT}}_{j+1} - \text{cycle}^{\mathrm{RPT}}_j} \times (Q^{\mathrm{RPT}}_{j+1} - Q^{\mathrm{RPT}}_j) + Q^{\mathrm{RPT}}_j
\end{equation}
where $\text{cycle}^{\mathrm{RPT}}_j$ and $\text{cycle}^{\mathrm{RPT}}_{j+1}$ denote the numbers of the aging cycle preceeding the $j$-th and $(j+1)$-th RPTs, respectively, while $Q^{\mathrm{RPT}}_j$ and $Q^{\mathrm{RPT}}_{j+1}$ represent the capacity values measured during these tests for the considered cell. Index $i$ ranges from 1 to the number of aging cycles a cell has undergone (Table \ref{tab:cells}, fourth column), while index $j$ ranges from 1 to the number of times the cell has been tested (Table \ref{tab:cells}, third column).

For example, capacity for cell V4 at  aging cycle \#30, namely $Q_{30}^{\mathrm{V4}}$,  is defined as:
\begin{equation}
	Q_{30}^{\mathrm{V4}} = \frac{30 - \text{cycle}^{\mathrm{RPT,V4}}_2}{\text{cycle}^{\mathrm{RPT,V4}}_{3} - \text{cycle}^{\mathrm{RPT,V4}}_2} \times (Q^{\mathrm{RPT,V4}}_{3} - Q^{\mathrm{RPT,V4}}_2) + Q^{\mathrm{RPT,V4}}_2
\end{equation}
where $\text{cycle}^{\mathrm{RPT}}_2 = 20$ and $\text{cycle}^{\mathrm{RPT}}_{3} = 45$, since cell V4 has undergone 20 aging cycles before RPT \#2 and 45 aging cycles before RPT \#3.

\subsection*{Definition of SOH indicators}\label{sec:SOH_indicator_formulation}
$V_{\mathrm{ch}}$ and $V_{\mathrm{dis}}$ represent the voltage profiles during charging and discharging, respectively. $I_{\mathrm{ch}}$ and $I_{\mathrm{dis}}$, are the current profiles during charging and discharging, respectively.
Voltage variations due to acceleration peaks during discharging are indicated with $\Delta V_{\mathrm{acc}}$, and the corresponding current variations with $\Delta I_{\mathrm{acc}}$.
The autocorrelation function measures the linear relationship between a signal $x(t)$ and its time-delayed version $x(t+\tau)$, where $\tau$ is the time delay. In this work, power autocorrelation during the discharge phase is quantified by correlating the power signal with its delayed copies. First, cell power is calculated from the voltage and current signals as follows:
\begin{equation}
	P(t) = V_{\mathrm{dis}}(t) \cdot I_{\mathrm{dis}}(t)
\end{equation} 
The autocorrelation function of the power signal $\hat{\rho}_{\tau}$ is computed with delays $\tau$ limited to a range  $[-\tau_{max}, \tau_{max}]$. In our study, $\tau_{max}$ is set to 3000 seconds. For each value within this range, $\hat{\rho}_{\tau}$  is computed as follows:
\begin{equation}
	\hat{\rho}_{\tau} = \sum_{t=\tau+1}^{T}(P(t)-\bar{P})(P(t-\tau)-\bar{P})
	\label{eq:Autocorrelation}
\end{equation}
where $T$ is the duration of the discharging phase, $P(t)$ is the power at time $t$, $\bar{P}$ is the average of the power over the time window $T$, and $P(t-\tau)$ is the power at instant $t-\tau$. The power autocorrelation indicator $P_{\mathrm{Autocorr}}$ is defined as the autocorrelation with null delay: $P_{\mathrm{Autocorr}}=\hat{\rho}_{\tau=0}$.

The resistance $R$ indicator is extracted for each aging cycle during the discharging phase using the following procedure. 
First, acceleration peaks are identified during the discharge\cite{allamnc} and depicted in Supplementary Figure S\ref{fig:peakcomputation}. Then, the resistance $R_{peak}$ corresponding to the $l^{\mathrm{th}}$ current peak within the $i^{\mathrm{th}}$ aging cycle is computed as follows:
\begin{equation}\label{eq:R_peak}
	R_{\mathrm{peak},l}^i = \frac{\Delta V_{\mathrm{acc},l}^i}{\Delta I_{\mathrm{acc},l}^i}
\end{equation}
where $\Delta V_j^i$ and $\Delta I_j^i$ are the voltage and current variations at the peak occurrence, respectively, as shown in Supplementary Figure S\ref{fig:peakcomputation}.
Thus, $P$ resistances $R_{\mathrm{peak},1}^i, R_{\mathrm{peak},2}^{i}, ..., R_{\mathrm{peak},P}^{i}$ are computed 
for each $i^{th}$ aging cycle, with $i=1, ..., N$, where $N$ represents the number of aging cycles during the cell's life and $P$ is the total number of acceleration peaks within each cycle.  Note that the number of total accelatrion peaks, $P$, varies with the aging cycle.
Subsequently, a single resistance value for each aging cycle is obtained by averaging the  $P$  resistances extracted from all acceleration peaks within that cycle:
\begin{equation}
	R^{i} = \frac{\sum_{l=1}^{P} R_{\mathrm{peak},l}^{i}}{P} ~~~~i=1,2,...,N
\end{equation}

The instantaneous battery charging impedance $Z_{\mathrm{CHG_{ist}}}$  is computed over the CC-a phase\cite{allamnc} as follows:
\begin{equation}
	Z_{\mathrm{CHG_{ist}}}(t_k) = -\frac{V_{\mathrm{ch}}(t_{k})-V_{\mathrm{ch}}(t_{k-1})}{I_{\mathrm{ch}}}
	\label{eq:imp_1_ist}
\end{equation}
where $V_{\mathrm{ch}}(t_{k}) - V_{\mathrm{ch}}(t_{k-1})$ is the voltage difference over the interval $\Delta t = t_{k} -t_{k-1}$, and $I_{\mathrm{ch}}$ is the constant charging current during the CC-a phase.

The choice of the time window $\Delta t$ is crucial. Increasing $\Delta t$ helps filter out  noise from the voltage difference in the numerator of Equation \eqref{eq:imp_1}  and reduces current quantization effects. 
However, too large a window can excessively filter and result in information loss. 
Therefore,  $\Delta t$  is tuned to balance noise reduction while preserving the information content of $Z_{\mathrm{CHG_{ist}}}$. 
The time intervals $\Delta t$ are selected based on the \textit{C}-rate: $\Delta t = 60\,s$ for \textit{C}/4, $\Delta t = 30\,s$ for \textit{C}/2, and $\Delta t = 1\,s$ for 1\textit{C} charging events.

After extracting the instantaneous battery impedance for all the time intervals of the charging phase, the $Z_{\mathrm{CHG}}$ indicator is computed for each charging phase by averaging the $Z_{\mathrm{CHG_{ist}}}$ within a specific voltage range $[V_{\mathrm{in}}, V_{\mathrm{fin}}]$:
\begin{equation}
	Z_{\mathrm{CHG}} = \frac{1}{M} \sum_{t_k=t_{\mathrm{in}}}^{t_{\mathrm{fin}}} Z_{\mathrm{CHG_{ist}}}(t_k)
	\label{eq:imp_1}
\end{equation}
where $M$ is the number of $Z_{\mathrm{CHG_{ist}}}$ measurements within  the considered voltage range, and $t_{\mathrm{in}}$ and $t_{\mathrm{fin}}$ are the initial and final time instants such that $V(t_{\mathrm{in}}) = V_{\mathrm{in}}$ and $V(t_{\mathrm{fin}}) = V_{\mathrm{fin}}$,  respectively.
The voltages $V_{\mathrm{in}}$ and $V_{\mathrm{fin}}$ were set to 3.8 V and 3.9 V, respectively, based on the sensitivity analysis presented in the Supplementary Note \nameref{sup:voltage_sensitivity_impedance}.

An alternative formulation would be to compute the average of $Z_{\mathrm{CHG_{ist}}}$ within a SOC range instead of a voltage range.
However, we opted for the voltage-based formulation to avoid estimation errors affecting the SOC, which is a non-measurable quantity generally estimated by the BMS. Additionally, a different definition of charging impedance, discussed in Supplementary Note \nameref{sup:impedance_with_OCV}, has been excluded in the present work due to its lower correlation with capacity fade.

Finally, the energy during charging and discharging is computed on the CC-a charging segment (see Fig. \ref{fig:OCV}) and driving UDDS profile, respectively, by integrating the electrical power within a fixed voltage window, specifically $[V_{\mathrm{in,ch}}, V_{\mathrm{fin,ch}}]$ and $[V_{\mathrm{in,dis}}, V_{\mathrm{fin,dis}}]$:
\begin{flalign}
& E_{\mathrm{ch}} = \int_{t_{\mathrm{in}}}^{t_{\mathrm{fin}}}V_{\mathrm{ch}}(t)\cdot I_{\mathrm{ch}}(t)\ \text{dt} \\
& E_{\mathrm{dis}} = \int_{t_{\mathrm{in}}}^{t_{\mathrm{fin}}}V_{\mathrm{dis}}(t)\cdot I_{\mathrm{dis}}(t)\ \text{dt}
\end{flalign}
where $V_{\mathrm{ch}}$ is the cell voltage during charging,  $I_{\mathrm{ch}}$ is the cell current during charging and $t_{\mathrm{in}}$ and $t_{\mathrm{fin}}$ are the initial and final time instants such that $V_{\mathrm{ch}}(t_{\mathrm{in}}) = V_{\mathrm{in,ch}}$ and $V_{\mathrm{ch}}(t_{\mathrm{fin}}) = V_{\mathrm{fin,ch}}$. Similarly, $V_{\mathrm{dis}}$ is the cell voltage during discharging,  $I_{\mathrm{dis}}$ is the cell current during discharging and $t_{\mathrm{in}}$ and $t_{\mathrm{fin}}$ are the initial and final time instants such that $V_{\mathrm{dis}}(t_{\mathrm{in}}) = V_{\mathrm{in,dis}}$ and $V_{\mathrm{dis}}(t_{\mathrm{fin}}) = V_{\mathrm{fin,dis}}$.  
Thus, energy is not only a function of the \textit{C}-rate but also depends on the voltage window over which it is calculated.

We selected the fixed voltage windows [$V_{\mathrm{in,ch}}$ = 3.6 V, $V_{\mathrm{fin,ch}}$ = 3.9 V] and [$V_{\mathrm{in,dis}}$ = 3.85 V, $V_{\mathrm{fin,dis}}$ = 3.4 V] for computing $E_{\mathrm{ch}}$ and $E_{\mathrm{dis}}$, respectively, to bypass the initial and final stages of charging and discharging, which are potentially prone to noise.

% To ensure consistency and comparability throughout the battery's life, $E_{ch}$ needs to be computed during constant current charging and over a fixed voltage window $[V_i,V_f]$, where $V_i$ and $V_f$ refer to the lower and upper limits of the voltage window, respectively. 
% This particular choice ensures that $E_{ch}$ values can be consistently compared across the same voltage range, regardless of the \textit{C}-rates at which the cells are charged.
% In this work, $E_{ch}$ is computed within the voltage range [3.6 V, 3.9 V], which falls within the CC-a phase. 

\subsection*{ Sensitivity of Charging Energy to Voltage Window}
To assess the feasibility of using energy during charging for partial charging profiles, the correlation between
 $E_{\mathrm{ch}}$ and capacity loss was quantified across different voltage ranges.
First, the interval [$V_{\mathrm{in,ch}}$, $V_{\mathrm{f,ch}}$] was divided into sub-intervals of 0.25 V amplitude, and the energy was computed for each sub-interval.

\begin{figure}[H]
	\centering
	\includegraphics[width = 1\textwidth]{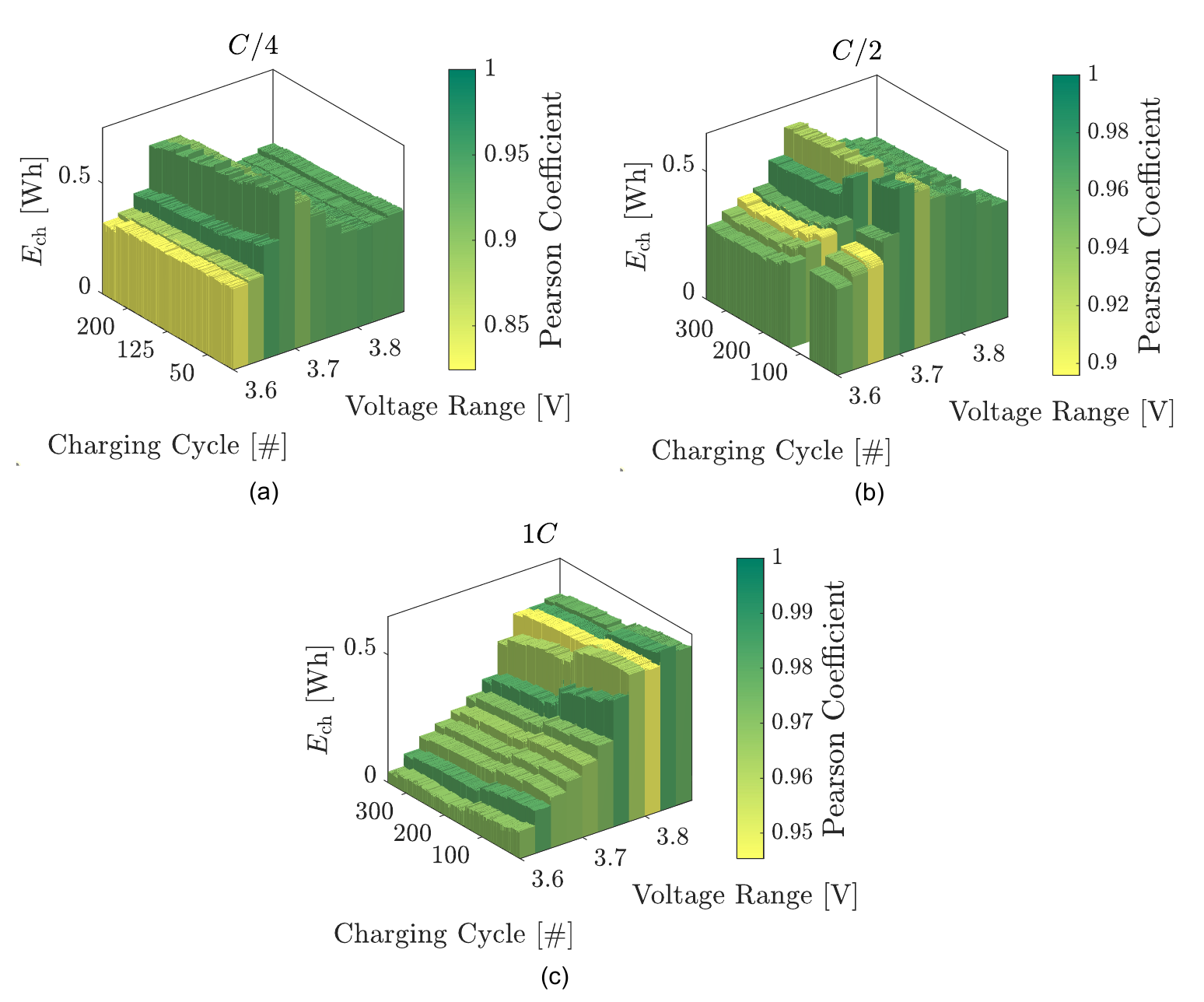}
	\caption{}
	% \caption{\textbf{Impact of Voltage Range and C-rate on Energy during charging}. The amount of energy during charging \rev{($E_\mathrm{ch}$)} depends on the voltage range  and the charging rate. As the charging rate increases (from  \textit{C}/4 for cell V4 (\textbf{a}), to  \textit{C}/2 for cell W8 (\textbf{b}), to  1\textit{C} for W9 (\textbf{c})), the peak in charging energy shifts towards higher voltage ranges, i.e., [3.675 V - 3.7 V] at \textit{C}/4,  [3.7 V - 3.725 V] at \textit{C}/2 and  [3.825 V - 3.85 V] at 1\textit{C}.}
	\label{fig:ChgEnergy_VoltageWindow} 
\end{figure}

As shown in Fig. \ref{fig:ChgEnergy_VoltageWindow}, there is a strong correlation between energy during charging and capacity loss across all voltage sub-intervals.  These results show that energy can be effectively used to estimate the SOH for partial and narrow charging periods. The analysis  indicates that the voltage interval with the highest correlation also depends on the charging rate. This insight facilitates straightforward integration into the BMS.

\subsection*{Pre-processing and incremental indicators} \label{sec:incremental_features}
Data pre-processing is essential for effectively using SOH indicators in data-driven algorithms. A critical step is removing outliers—data points that deviate from the majority. Outliers can affect feature extraction and machine learning model performance. Therefore, a careful approach is used to remove outlier-containing data, ensuring more robust and reliable feature representation.
A second step of the pre-processing phase is the computation of the incremental features,  denoted by $\Delta$. 
This subsection explains how to obtain these features, using incremental resistances as an example.  

For each cell in the dataset, the vector of incremental resistances  $\Delta \textbf{R}$ is calculated as follows:
\begin{enumerate}
	\item For each aging cycle $i^{th}, i=1, ..., N$, the resistance during the discharge phase over acceleration peaks is calculated as a function of SOC.  $R^1$ represents  the average resistance over the SOC range of  80\% and 20\% during discharge.   The resulting resistance vector is: 
	\begin{equation}
		\textbf{R} = [R^1, R^2, ..., R^N]
	\end{equation}
	 where $R^1$ is the average fresh cell resistance and  $R^N$ is the average resistance at the last cycle. 
	\item Obtain the incremental resistance vector by subtracting $R^1$ from each value in $R$. 
	\begin{equation}
		\Delta \textbf{R} = \textbf{R} - R^1
	\end{equation}
\end{enumerate}

This approach ensures that the first element of the incremental vector for each feature is zero, facilitating comparison of aging trends across cells. Additionally, the charging impedance vector $\textbf{Z}_{\mathrm{CHG}}$ requires further pre-processing due to its dependency on the \textit{C}-rate (see,  Fig. \ref{fig:pR_withandwithoutOCV}.  this feature strongly depends on the \textit{C}-rate at which it is computed. 
To standardize across different \textit{C}-rates, the incremental vector $\Delta \textbf{Z}_{\mathrm{CHG}}$  is normalized using the fresh cell impedance value: 
\begin{equation}
    \Delta \textbf{Z}_{\mathrm{CHG}}^{\mathrm{NORM}} = \frac{\Delta \textbf{Z}_{\mathrm{CHG}}}{Z_{\mathrm{CHG}}^1}
\end{equation}
where $Z_{\mathrm{CHG}}^1$ is the charging impedance calculated over the first aging cycle in Batch \#1 in the voltage range [3.8 V - 3.9 V] as described in Sec. \nameref{sec:SOH_indicator_formulation}.
Normalization reduces variations from different charging rates, providing a consistent feature representation. This pre-processing step is crucial for evaluating the ML model across cells cycled at various rates, effectively excluding \textit{C}-rate as a training feature.  It ensures a more refined data representation for machine learning algorithms.

\subsection*{Estimation model}
In this work, the LRM  estimates capacity fade due to its strong linear correlation with SOH indicators.
The LRM relates the response variable $y$ to the input vector $\boldsymbol{u}$ as follows\cite{montgomery2021introduction}:
\begin{equation}
	y(t) = \beta_0 + \beta \boldsymbol{u}(t) + \epsilon(t)
	\label{eq:LRM}
\end{equation}
where $\epsilon$ represents model error, capturing deviations between the model and observed data. Coefficients $\beta$ are determined using the least-squares method, which minimizes the model error on the training dataset.
To evaluate the accuracy of the estimation models, the root mean square error (RMSE) is calculated as:
\begin{align} \label{eq:RMSE}
	\mathrm{RMSE} &= \sqrt{\frac{\sum_{i=1}^N e_i^2}{N}} \\
	e_i &= \frac{Q_{\mathrm{cell},i} - Q_{\mathrm{est},i}}{Q_{\mathrm{cell},i}}
\end{align}
where $e_i$ is the relative error,  with $Q_{\mathrm{cell},i}$ and $Q_{\mathrm{est},i}$ representing the actual  and estimated capacities  at the cycle $i$, respectively.
Additionally, the absolute percentage error  is given by $\mathrm{APE} (\%)=|e_i| \cdot 100$.

\section*{Data Availability Statement}
All experimental data\cite{pozzato2022lithium} are available online at the following Open Science Framework repository:
\href{https://osf.io/qsabn/?view_only=2a03b6c78ef14922a3e244f3d549de78}{OSF}.
%The dataset collects experimental data for ten INR21700-M50T battery cells with graphite/silicon anode and NMC cathode tested in a temperature controlled chamber at 23$^\circ$C.  The UDDS driving cycle was used to cycle and age the cells,  while periodic RPTs,  in the form of HPPC,  EIS,  and \textit{C}/20 capacity test,  were performed to assess the battery health status.
% DV curves and hysteresis at different temperatures (Note S\ref{sup:c}) are computed from charge and discharge OCVs acquired from two pristine INR21700-M50T battery cells.  Cells are CC charged and discharged at \textit{C}/40 considering three different temperatures,  namely 10,  23,  and 40$^\circ$C.  The dataset is made available to the public at the following Mendeley Data repository: \href{https://data.mendeley.com/datasets/v85vp6xwzh/draft?a=4a34735a-c4b7-433e-9763-9e8cb867e2a9}{Mendeley Data}. 

\section*{Code Availability Statement}
The code supporting the findings of this study is available at the following Open Science Framework repository: \href{https://osf.io/347ep/?view_only=3c79745a646a4fcd92b4c955b4a99f24}{OSF}.

\bibliography{sn-bibliography}% common bib file

\section*{Acknowledgements}
The authors thank the members and alumni of the Stanford Energy Control Lab: Luca Pulvirenti,  Sara Ha, Sai Thatipamula, Muhammad Aadil Khan and, in particular, Le Xu for their feedback on the manuscript.
This work was partially funded by the Precourt Institute of Energy, Stanford University. This research is enabled in part through computational resources and support provided by Sherlock compute cluster of Stanford University.

\section*{Author Contributions}
Conceptualization: A.A., G.P., and S.O.; data curation: A.L., G.P., and P.B.; formal analysis: A.L., M.A., and P.B.; funding acquisition: S.O.; investigation: A.L., G.P., P.B., and S.O.;
methodology: A.L., G.P., M.A., and P.B., and S.O.; project administration: S.O.;
resources: S.O.; supervision: S.O.;
visualisation: A.L., G.P., M.A., and P.B.; validation:  A.L., M.A., and P.B.;
writing—original draft: A.L., G.P., M.A., P.B., S.O.; writing—review and editing:
A.L., M.A., P.B., S.O.

\section*{Competing interests}
The authors declare the following competing interests: A.A. is with Archer Aviation and G.P. is with Form Energy. They were both
affiliated with Stanford University at the time of the research.


\section*{Supplementary Information}
The online version contains
supplementary material available at

\section*{Figure Legends}
\begin{itemize}
	\item \textbf{Fig. 1:} \textbf{Power autocorrelation.} \textbf{a} Power autocorrelation ($P_{\mathrm{Autocorr}}$) profiles calculated over discharge profiles for cell W8 throughout  its lifetime. \textbf{b} Percentage decrease of the peak amplitude at null delay ($P_{\mathrm{Autocorr,loss}}$) is plotted against the percentage capacity loss ($Q_{\mathrm{cell,loss}}$).
	\item \textbf{Fig. 2:} \textbf{Resistance.} \textbf{a} Internal resistance ($R$) is plotted as a function of the current peak number during the discharge phases throughout the cell’s life span for cell W8. Shades of gray represent different batches of aging cycles. Batch $j$ is defined as the period between  the $j$th  and the $(j+1)$th RPT. Green points indicate the average internal resistance. \textbf{b} Percentage increase in internal resistance ($R_{\mathrm{increase}}$) relative to capacity loss ($Q_{\mathrm{cell,loss}}$) for all five cells.
	\item \textbf{Fig. 3:} \textbf{Charging Impedance ($\mathbf{Z_{\mathrm{CHG}}}$) as function of SOC and cycle number for cells V4 (a), W8 (b), and W9 (c).} The yellow curve represents fresh cell conditions, while the dark blue curve denotes aged cell conditions.
	\item \textbf{Fig. 4:} \textbf{Charging Impedance.} \textbf{a} Charging Impedance ($Z_{\mathrm{CHG}}$) as a function of voltage for cell W8. Charging impedance is averaged over the voltage range [$V_{\mathrm{in}}$ = 3.8 V, $V_{\mathrm{fin}}$ = 3.9 V]. \textbf{b} Percentage variation of average $Z_{\mathrm{CHG}}$ ($Z_{\mathrm{CHG,increase}}$) as a function of capacity fade ($Q_{\mathrm{cell,loss}}$).
	\item \textbf{Fig. 5:} \textbf{Energy during charging and discharging.} \textbf{a}  Energy during charging ($E_{\mathrm{ch}}$) as a function of charging time within the voltage range [$V_{\mathrm{in,ch}}$ = 3.6 V,$V_{\mathrm{fin,ch}}$ = 3.9 V] for cell W8 throughout its cycle life. \textbf{b}  Energy loss during charging ($E_{\mathrm{ch,loss}}$) across all 5 cells shows a linear correlation with capacity loss ($Q_{\mathrm{cell,loss}}$). \textbf{c} Energy during discharging ($E_{\mathrm{dis}}$) as a function of discharging time within the voltage range [$V_{\mathrm{in,dis}}$ = 3.85 V, $V_{\mathrm{fin,dis}}$ = 3.4 V] for cell W8 throughout its cycle life. \textbf{d}  Energy loss during discharging ($E_{\mathrm{dis,loss}}$) for all five cells demonstrates a linear correlation with capacity loss ($Q_{\mathrm{cell,loss}}$).
	\item \textbf{Fig. 6:} \textbf{Correlation Analysis.} \textbf{a} Heatmap showing the Pearson's correlation coefficients between incremental power autocorrelation ($\Delta \textit{P}_{\mathrm{Autocorr}}$), incremental energy during charging ($\Delta \textit{E}_{\mathrm{ch}}$), incremental energy during discharging ($\Delta \textit{E}_{\mathrm{dis}}$), normalized incremental charging impedance ($\Delta \textit{Z}_{\mathrm{CHG}}^{\mathrm{NORM}}$), and incremental resistance ($\Delta R$) with capacity loss for each individual cell.  \textbf{b} Histogram illustrating the Pearson's correlation coefficients between incremental features and capacity loss across all cell data. Note that the correlation between the capacity of cell W7 and $\Delta R$ is not reported due to some computed resistances being deemed unreliable because of data acquisition issues  (Sec. \nameref{sec: dataset} and Supplementary Note \nameref{sup:IncreasedVoltage}), and thus interpreted as outliers during the pre-processing phase.
	\item \textbf{Fig. 7:} \textbf{SOH estimation results from the LRM using power autocorrelation ($\mathrm{LRM}_{P_\mathrm{Autocorr}}$), and energy during charging and discharging ($\mathrm{LRM}_{E_\mathrm{ch},E_\mathrm{dis}}$) as input features.} Profiles of capacity loss and estimation error for cells  V4 (\textbf{a}),  W5 (\textbf{b}),  W7 (\textbf{c}), and W9 (\textbf{d}). Augmented capacity points (obtained as discussed in Sec. \nameref{sec:data_augmentation}) are shown in red. SOH estimation using power autocorrelation as input is shown in brown (with training data from cell W8) and yellow (with training data from all cells except the test cell). The dark blue and light blue lines show SOH estimation using energy features as input, with training data from cell W8 (dark blue) and from all available cells except the test cell (light blue). Gaps in the capacity curves for cells W5 (\textbf{b}) and W7 (\textbf{c}) are due to voltage measurements anomalies affecting the reliability of feature values (see Sec. \nameref{sec: dataset} and Supplementary Note \nameref{sup:IncreasedVoltage}). The capacity drop for cell W8 (\textbf{d}) results from issues with the aging protocol implementation.
	\item \textbf{Fig. 8:} \textbf{SOH estimation results from the LRM using charge and discharge energies, charging impedance, and resistance as features.}
	The capacity loss and estimation error profiles for cells V4 (\textbf{a}), W5 (\textbf{b}), W7 (\textbf{c}), and W9 (\textbf{d}) are shown. Augmented Capacity points (obtained as discussed in Sec. \nameref{sec:data_augmentation}) are shown in red. Three scenarios are displayed: blue represents the LRM output trained solely with energy during charging and energy during discharging ($\mathrm{LRM}_{E_\mathrm{ch},E_\mathrm{dis}}$); the light purple represents the LRM output trained with energy during charging, energy during discharging along with charging impedance ($\mathrm{LRM}_{E_\mathrm{ch},E_\mathrm{dis},Z_\mathrm{CHG}}$); the dark purple represents the LRM output trained with energy during charging, energy during discharging, charging impedance, and resistance ($\mathrm{LRM}_{E_\mathrm{ch},E_\mathrm{dis},Z_\mathrm{CHG},R}$). All models are trained exclusively using data from cell W8.
	\item \textbf{Fig. 9 :} \textbf{Impact of Voltage Range and $C$-rate on Energy during charging}. The amount of energy during charging ($E_\mathrm{ch}$) depends on the voltage range  and the charging rate. As the charging rate increases (from  \textit{C}/4 for cell V4 (\textbf{a}), to  \textit{C}/2 for cell W8 (\textbf{b}), to  1\textit{C} for W9 (\textbf{c})), the peak in charging energy shifts towards higher voltage ranges, i.e., [3.675 V - 3.7 V] at \textit{C}/4,  [3.7 V - 3.725 V] at \textit{C}/2 and  [3.825 V - 3.85 V] at 1\textit{C}.
\end{itemize}

\section*{Tables}

\begin{table}[ht] 
	\centering
	\normalsize
	\begin{tabular}{ccccc}
			\textbf{Cell identifier} & \begin{tabular}{c} \textbf{CC-a} [1/h] \end{tabular} & \textbf{RPTs [\#]} & \textbf{Cycles [\#]} \\ \hline
			Cell V4 & \textit{C}/4  & 10 & 244\\
			Cell W5 & \textit{C}/2  & 14 & 369\\
			Cell W7 & \textit{C}/4  & 5 & 141\\
			Cell W8 & \textit{C}/2  & 14 & 347\\
   			Cell W9  & 1\textit{C}  & 14 & 341\\
			\hline 
	\end{tabular}\vspace{0.5em}
    \caption{\textbf{Battery cells.} Cells identifiers,  charge \textit{C}-rate, total number of RPTs and total number aging cycles of the battery cells\cite{pozzato2022lithium} used in this work. Tests were conducted at a controlled temperature of $23\mathrm{^\circ C}$. See Supplementary Note \nameref{sup:experimental_data} for further information on the dataset. }\label{tab:cells}
\end{table}

%%%%%%%%%%%%%%% SUPPLEMENTARY MATERIAL
\setcounter{figure}{0}  
\setcounter{table}{0}  
\setcounter{equation}{0}  
\setcounter{section}{0}  
\setcounter{section}{0}  
\setcounter{footnote}{0}  
%%%%%%%%%%%%%%% SUPPLEMENTARY MATERIAL 
\makeatletter
\renewcommand{\fnum@figure}{Supplementary Figure S\thefigure}
\makeatother
\makeatletter
\renewcommand{\fnum@table}{Supplementary Table S\thetable}
\makeatother
\renewcommand{\theequation}{S.\arabic{equation}}
\newpage
\clearpage
\begin{titlepage}
    \centering
    \vspace*{\fill}

    \vspace*{0.5cm}
    \huge
    \textbf{Supplementary Information}

    \vspace*{0.5cm}
    \large State of health indicators for battery capacity,  energy,  and power fade quantification

    \vspace*{\fill}
\end{titlepage}

\clearpage

\section*{Supplementary Notes}
\subsection*{Machine-learning pipeline} \label{sup:MLpipeline}
Data-driven model development for SOH estimation comprises of four stages: feature engineering, data pre-processing, model
development, and model evaluation, as depicted in Supplementary Figure S\ref{fig:ml_pipeline}.
The feature engineering phase starts from the structured dataset and involves calculating the SOH indicators using current and voltage signals. Subsequently, outliers are removed and features are normalized/standardized to increase their generalizability over the entire set of cells cycled at different \textit{C}-rates. 
Based on their correlations with capacity and on their mutual correlation, the most indicative features are selected as input to the estimation models. 
After the dataset is partitioned into training and testing, the model development phase involves the training of regression models like FNN, RNN, LRM and ARMAX.
%which can include iterative processes such as the optimization of the model hyperparameters. 
Finally, the model is tested to evaluate its estimation accuracy on the test portion of the dataset, which is different from the one used for training.
\begin{figure}[h]
\centering
\includegraphics[width = 1.2\textwidth]{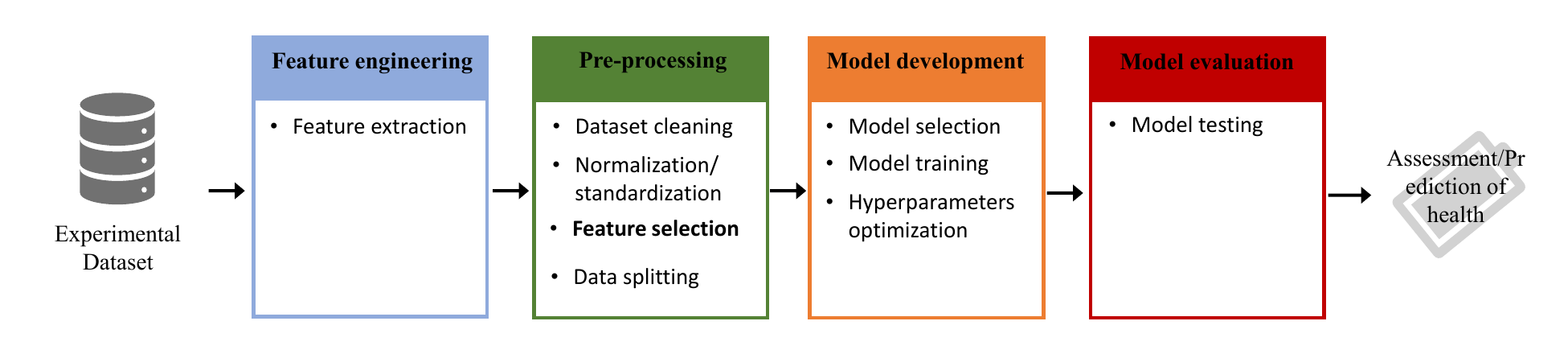}
\caption{\textbf{Machine learning pipeline.} First, features are extracted from experimental data, and then the most indicative features are selected as inputs to the ML model. 
Different machine learning models are trained and optimized, and their performance evaluated for SOH estimation
to identify the best model in terms of accuracy and computational time.}
\label{fig:ml_pipeline} 
\end{figure}

\clearpage

\subsection*{Resistance during the discharge phase over acceleration peaks} \label{sup:resistance}

\begin{figure}[H]
	\centering
	\includegraphics[width = 1\textwidth]{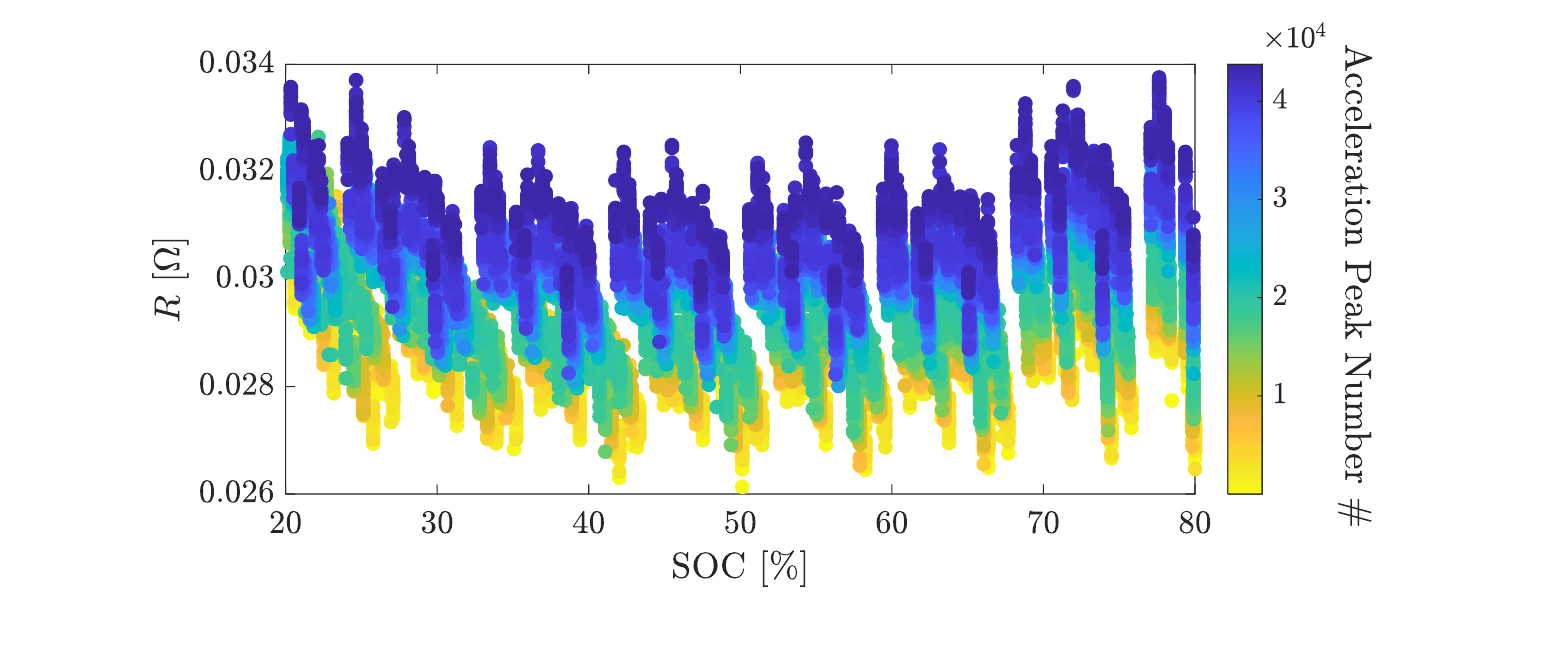}
	\caption{\textbf{Resistance ($R$) during the discharge phase over acceleration peaks. }Resistances of cell W8 calculated during acceleration peaks, from the fresh cell (yellow) to the aged cell (dark blue) as a function of SOC in the  range from $20\%$ to $80\%$.}
	\label{fig:ResistanceSOC} 
\end{figure}

%To ensure that the resistance $R$ calculated in the discharge phase over acceleration peaks reflects only the battery's health, its relationship with SOC was assessed. 
Figure S\ref{fig:ResistanceSOC} presents the resistance values of cell W8 over its lifespan as a function of SOC. The color gradient in the plot indicates the cell's age, from fresh (yellow) to aged (dark blue).  This resistance is averaged over the $SOC$ range of 80\% to 20\% when evaluating the internal resistance feature vector.

%\rev{ Figure S\ref{fig:DeltaResistanceSOC} shows the difference in internal resistance between the first and last resistance values computed at a specific SOC.  Although resistance values clearly increase as the cell ages,no specific SOC range exhibits a more pronounced increase of resistance.}

%\rev{\begin{figure}
%	\centering
%	\includegraphics[width = 1\textwidth]{DeltaR_SOC.pdf}
%	\caption{\textbf{Resistance variation as function of SOC. } For Cell W8, The SOC range from $20\%$ to $80\%$ is divided into intervals with an amplitude of $2\%$, and the difference between the first and last computed resistance  within each interval is evaluated.}
%	\label{fig:DeltaResistanceSOC} 
%\end{figure}}

\clearpage

\subsection*{Charging Impedance with OCV} \label{sup:impedance_with_OCV}
Another possible formulation of the impedance indicator is the charging impedance with OCV, $Z_{\mathrm{CHG}_2}$, which relies on the zero-order equivalent circuit model of the battery cell.
 As the name suggests, the OCV curve is required to compute  $Z_{\mathrm{CHG}_2}$.
In this work, the battery pseudo-OCV is obtained through \textit{C}/40 experiments, performed  at controlled temperature of 23$\mathrm{^\circ C}$. As shown in Supplementary Figure S\ref{fig:OCV}, the hysteresis phenomenon results in two different SOC-OCV curves during the charge and discharge \textit{C}/40 tests.

\begin{figure}[H]
	\centering
	\includegraphics[width = 0.8\textwidth]{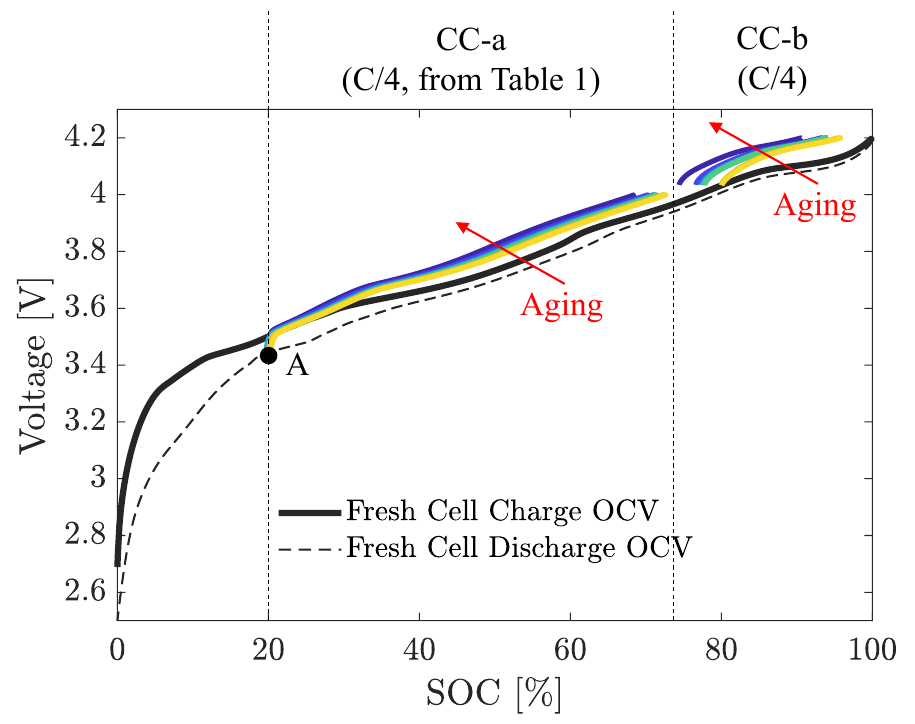}
	\caption{\textbf{Charge and discharge OCV for cell V4.} The battery voltage curves during CC-a and CC-b constant current phases are shown. As the battery ages, charging curves shift from right to left, with the yellow curve representing  fresh condition and the dark blue curve indicating the most aged state. At point \textbf{A}, due to the polarization from the previous discharge, the voltage at the start of charging aligns with the fresh cell's discharge OCV. Impedance calculated at this point can become negative as the battery's voltage falls below the charge OCV. This typically occurs at the start of charging (around 20\% SOC in this study) due to voltage hysteresis from the preceding UDDS discharge.}
	\label{fig:OCV} 
\end{figure}
Interestingly, the terminal voltage profiles during the CC-a charging phase for cell V4, shown using colored lines, shift from right to left as the battery ages. This shift is attributed to lithium loss, which can alter the battery's equilibrium potential.

\begin{figure}[H]
	\centering
	\includegraphics[width = 1\textwidth]{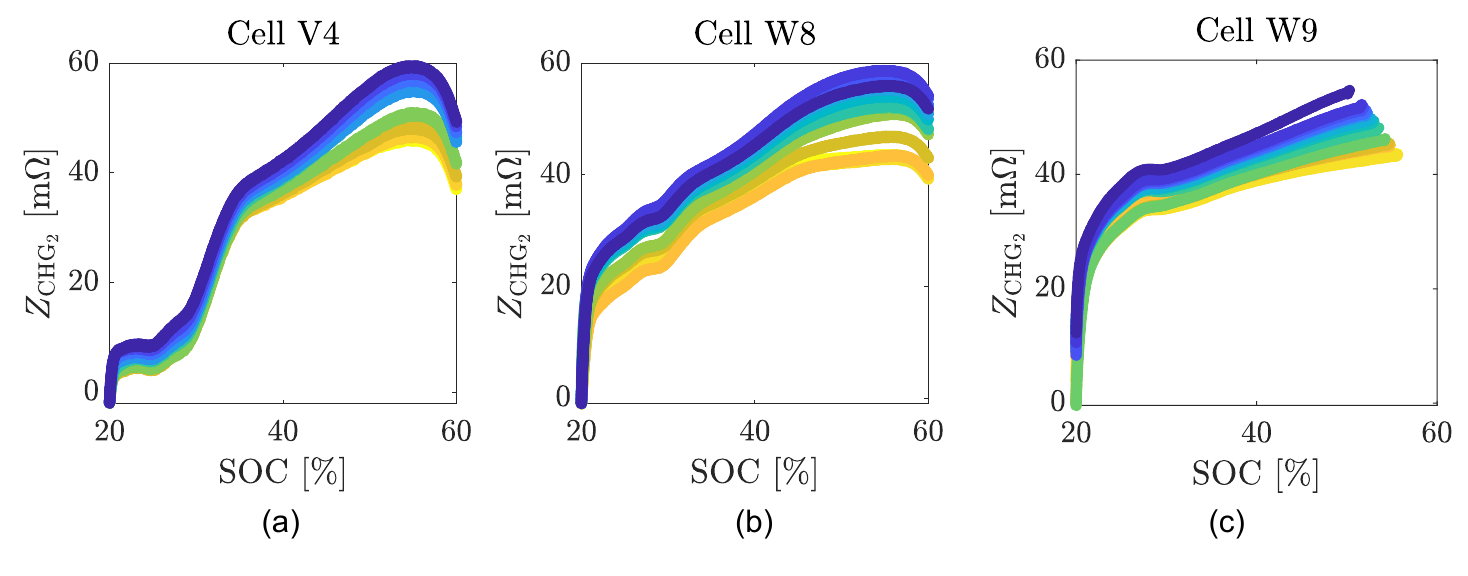}
	\caption{\textbf{Charging Impedance with OCV ($\mathbf{Z_{\mathrm{CHG}_2}}$) as function of SOC and cycle number for cells V4 (a), W8 (b), and W9 (c).} The yellow curve represents fresh cell conditions, while the dark blue curve denotes aged cell conditions.}
	\label{fig:pR_withOCV} 
\end{figure}

\begin{figure}[H]
	\centering
	\includegraphics[width = 1\textwidth]{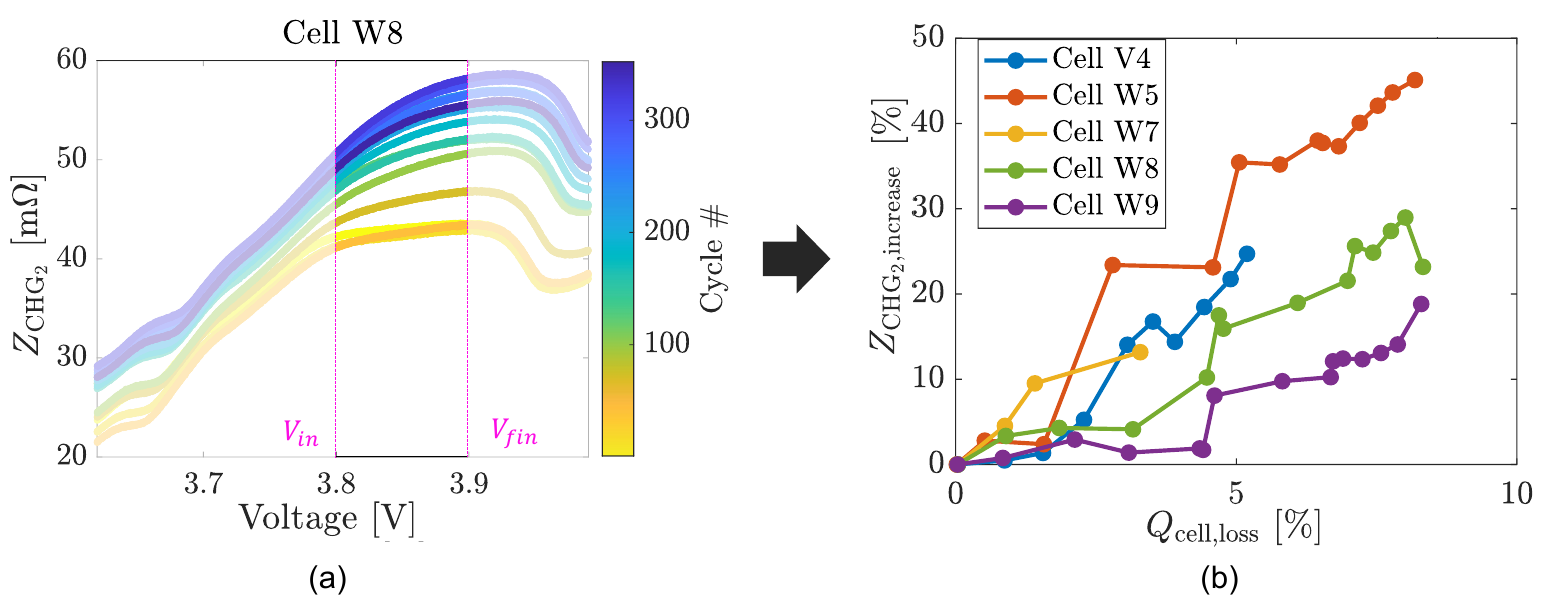}
	\caption{\textbf{Charging Impedance \textit{with} OCV.} \textbf{a} Charging Impedance \textit{with} OCV ($Z_{\mathrm{CHG}_2}$) as a function of voltage for cell W8. The voltage range [$V_{\mathrm{in}}$ = 3.8 V, $V_{\mathrm{fin}}$ = 3.9 V], over which the charging impedance \textit{with} OCV is averaged, is highlighted. \textbf{b} Variation of average $Z_{\mathrm{CHG}_2}$ as a function of capacity fade.}
	\label{fig:pR_withOCV_corr} 
\end{figure}

The impedance $Z_{\mathrm{CHG}_2}$ is determined during the constant current phase CC-a using the charge OCV through the following equation\cite{PletBook2}:
\begin{equation}
V(t) = V_{\mathrm{OCV}}(\mathrm{SOC}(t)) - Z_{\mathrm{CHG}_{2,\mathrm{ist}}}I_{dis}
\label{eq:zorder}
\end{equation}
Solving Equation \eqref{eq:zorder} for $Z_{\mathrm{CHG_{2,ist}}}$,  the following is obtained:
\begin{align}
	Z_{\mathrm{CHG_{2,ist}}}(t) &= -\frac{V(t)-V_{\mathrm{OCV}}(\mathrm{SOC}(t))}{I_{dis}}
\label{eq:imp_2}
\end{align}
where $V(t)$ is the voltage during charge,  $I_{dis}$ is the cell current (considered negative during charging),  and $V_{\mathrm{OCV}}$ is the fresh cell OCV in charge acquired at a \textit{C}-rate of \textit{C}/40.  As shown in Equation \eqref{eq:imp_2},  the OCV is a function of the SOC. For each aging cycle $i$, the impedance $Z_{\mathrm{CHG}_2}$ is defined as:
\begin{align}
	Z_{\mathrm{CHG}_2}^{i} &= \frac{1}{M} \sum_{t_k=t_{in}}^{t_{\mathrm{fin}}} Z_{\mathrm{CHG}_{2,\mathrm{ist}}}(t_k)
\label{eq:imp_3}
\end{align}

where $t_{\mathrm{in}}$ and $t_{\mathrm{fin}}$ are the initial and final time instants such that $V(t_{\mathrm{in}}) = V_{\mathrm{in}}$ and $V(t_{\mathrm{fin}}) = V_{\mathrm{fin}}$,  respectively. In this study, $Z_{\mathrm{CHG}}$ and $Z_{\mathrm{CHG_2}}$  are calculated within the same voltage window of $V_{\mathrm{in}}$ = 3.8 V and $V_{\mathrm{fin}}$ = 3.9 V. Supplementary Figure S\ref{fig:pR_withOCV} shows the $Z_{\mathrm{CHG_2}}$ extracted respectively for cell V4, W8 and W9. Supplementary Figure S\ref{fig:pR_withOCV_corr} describes the variations in the average impedance $Z_{\mathrm{CHG}_2}$, as a function of capacity fading. Unlike $Z_{\mathrm{CHG}}$, $Z_{\mathrm{CHG}_2}$ shows a low correlation with capacity loss.
As illustrated in Supplementary Figure S\ref{fig:OCV} (point \textbf{A}), the preceding UDDS discharge polarizes the battery toward the discharge OCV curve, which has lower values compared to the charging OCV curve. This results in a negative overvoltage ($V(t) - V_{\mathrm{OCV}} < 0$) and, consequently, a negative impedance at the start of CC-a charging. These negative impedance values are excluded from the analysis.

\begin{comment}

In this work,  indicators are computed considering a constant room temperature of 23$^\circ$C, and, hence, the OCV at 23$^\circ$C can be used to compute $Z_{CHG_2}$.    
However,  in real-world applications,  the behavior of the battery is a function of the environmental temperature\cite{allamnc}.   Tests performed in our laboratory show that increasing temperature from 10 to 40$^\circ$C leads to a modification of charge and discharge OCVs,  which in turn leads to a reduced hysteresis and to different lithium intercalation signatures (as shown in the supplemental Note S\ref{sup:c}).  Hence,  for a robust health assessment,  correlations between features and capacity fade over the battery lifetime should be established range of temperatures.  The analysis shown in Fig.s  \ref{fig:pR_withandwithoutOCV_corr},  and \ref{fig:energy} could be performed at different temperatures  to cover the whole range of battery operating conditions (usually defined in the battery cell technical specifications\cite{battechspec}).  Concerning $Z_{CHG}$,  fresh cell charge OCVs should be available at different temperatures and selected accordingly. 
Charge and discharge OCVs at different environmental temperatures are made available at the repository specified at the end of the paper. 

\end{comment}

\subsection*{Pseudo-DV curve}  \label{sup:pseudo-ocv}
$Z_{\mathrm{CHG}_2}$ can be interpreted through DV analysis\cite{allamnc}. The pseudo-DV $\mathrm{DV}{\mathrm{pseudo}}$ curve is calculated by scaling $Z_{\mathrm{CHG}_2}$ with $3600/\Delta t$:
       
\begin{equation}
    \mathrm{DV}_{\mathrm{pseudo}} = \frac{3600}{\Delta t}Z_{\mathrm{CHG}_2}
    \label{eq:dv}
\end{equation}
DV analysis evaluates electrochemical properties and aging effects by tracking changes in peaks and valleys\cite{dahn1991phase,winter1998insertion,pastor2017comparison}. It involves capacity tests at low \textit{C}-rates (typically $<$ $C$/10) to capture thermodynamic signatures and reduce  polarization and kinetic effects that increase with higher \textit{C}-rates\cite{fly2020rate}. DV curve computation is time-consuming and not always feasible.

In Supplementary Figure S\ref{fig:pR_withoutOCV_DV}, DV curves computed from \textit{C}/40 capacity test performed at 23$^\circ$C are compared with pseudo-DV curves derived from Equation \eqref{eq:dv}.  
Notably, the pseudo-DV curves effectively reproduce the key features of the DV curves at  \textit{C}/40 for both fresh and aged cells.  
This result demonstrates that CC charging at high \textit{C}-rate (\textit{C}/4, \textit{C}/2, and 1\textit{C}), can be used to reconstruct the local DV signature with reasonable accuracy.   Although increasing the \textit{C}-rate from \textit{C}/4 to 1\textit{C} results in smoother pseudo-DV curves, as shown in Supplementary Figure S\ref{fig:pR_withoutOCV_DV}, valleys at approximately 40\% and 90\% SOC are still discernible.

%While increasing the \textit{C}-rate from \textit{C}/4 to 1\textit{C} leads to smoother pseudo-DV curves,  as shown in Supplementary Figure S\ref{fig:pR_withoutOCV_DV},  valleys at  \raisebox{2.5pt}{\large\texttildelow}40\% and  \raisebox{2.5pt}{\large\texttildelow}90\% SOC can still be captured.
This finding eliminates the need for ongoing low \textit{C}-rate tests, which are impractical. A single \textit{C}/40 capacity test on a fresh cell can establish the full DV curve, capturing key features. Pseudo-DV can then track these features over the battery’s life and relate them to capacity fade. 
%, as shown in Fig. \ref{fig:pR_withoutOCV_corr} for $Z_{CHG_2}$
\begin{figure}[h]
\centering
\includegraphics[width = 1\textwidth]{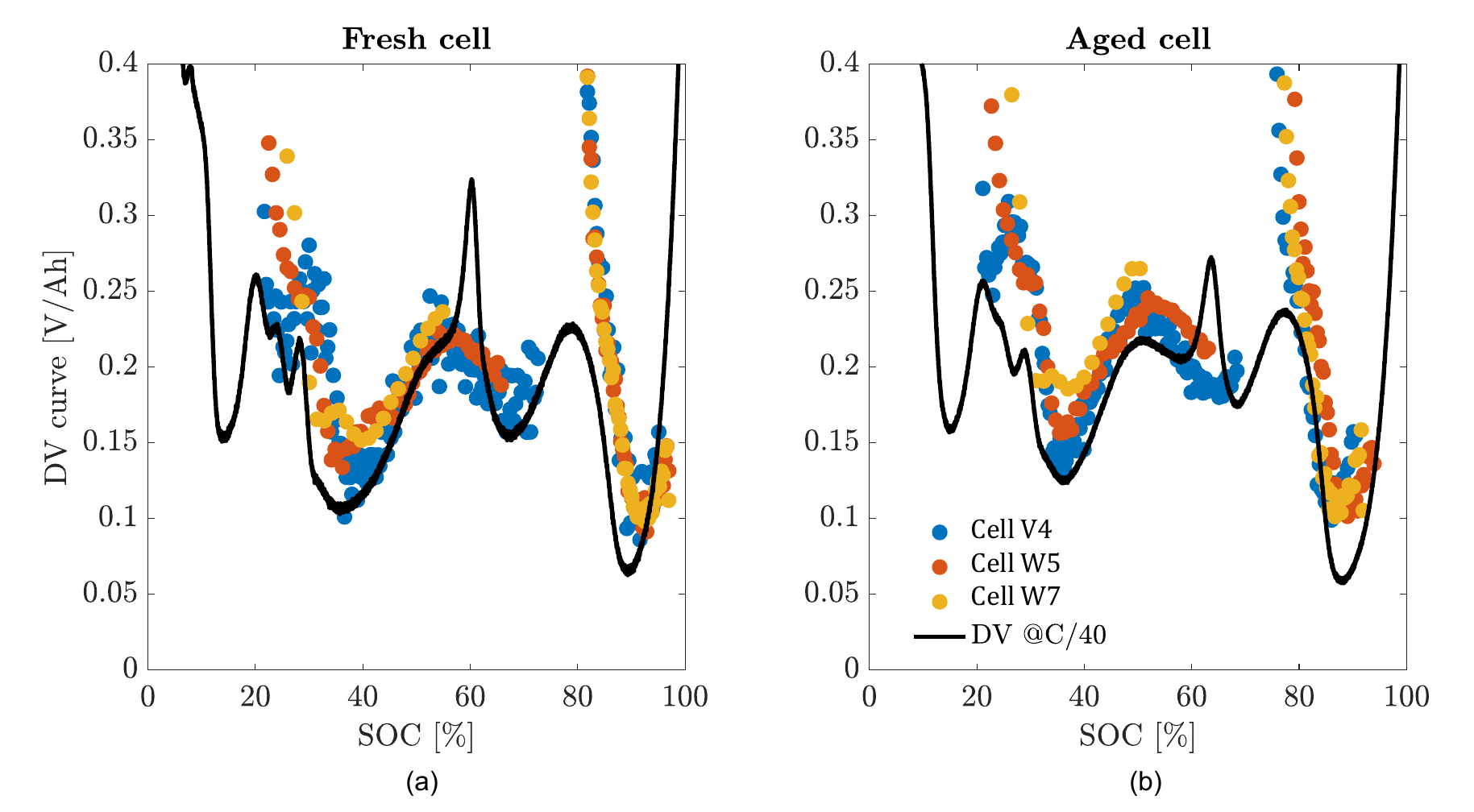}
\caption{\textbf{Pseudo-DV curves.} 
Comparison of DV voltage curves from \textit{C}/40 tests (black) with pseudo-DV curves derived from impedance $Z_{\mathrm{CHG}_2}$ (blue, red, and yellow) for cells V4, W5, and W7, charged at \textit{C}/4, \textit{C}/2, and 1\textit{C}, respectively. \textbf{(a)} shows curves for fresh cells, while \textbf{(b)} shows curves for aged cells. The results indicate that high \textit{C}-rate pseudo-DV curves can be used to approximate the \textit{C}/40 DV curves for both fresh and aged cells.}
\label{fig:pR_withoutOCV_DV} 
\end{figure}

\subsection*{Temperature effects on OCVs}\label{sup:t-ocv}
OCV curves are obtained by performing \textit{C}/40 experiments on a fresh cell at temperatures of 10, 23, and 40$^\circ$C. These curves are used to evaluate changes in hysteresis curves and differential voltage signatures as a function of temperature. The charge and discharge OCV experimental data were collected from two pristine INR21700-M50T battery cells.

\begin{figure}[H]
    \centering
    \includegraphics[width = 1\textwidth]{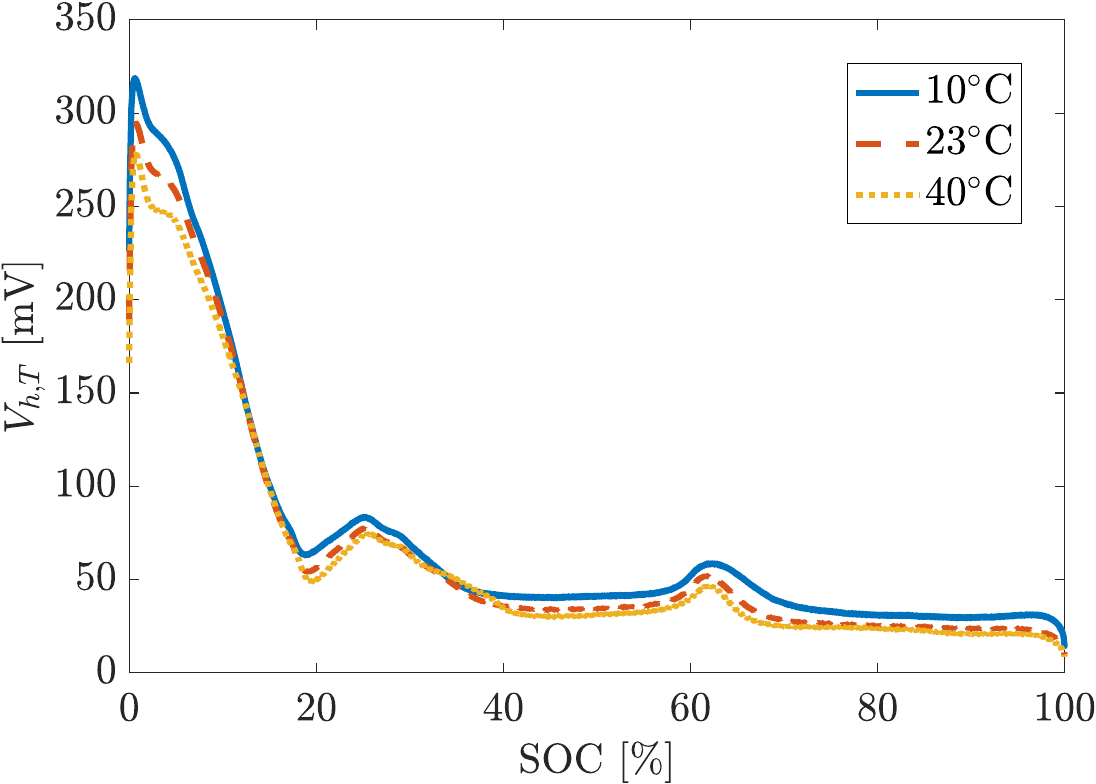}
    \caption{\textbf{OCV hysteresis.} OCV hysteresis ($V_{h,T}$) at 10, 23, and 40$^\circ$C as a function  of SOC.}
    \label{fig:hysttemp} 
\end{figure}

Supplementary Figure S\ref{fig:hysttemp} shows the OCV hysteresis computed as:
\begin{equation}
V_{h,T} = \mathrm{OCV}_{\mathrm{charge},T}-\mathrm{OCV}_{\mathrm{discharge},T}
\end{equation}
where $T$ is the experiment temperature (10,  23,  or 40$^\circ$C),  and $\mathrm{OCV}_{\mathrm{charge}}$ and $\mathrm{OCV}_{\mathrm{discharge}}$ are the charge and discharge OCV,  respectively.  

At SOC levels below 20\%, the voltage hysteresis ranges from 250 to 325 mV. This behavior, consistent with previous literature\cite{wycisk2022modified}, is attributed to silicon hysteresis in the negative electrode. Possible causes include the interaction between mechanical stresses and electric potential\cite{sethuraman2010situ}, and asymmetric reaction pathways during lithiation and delithiation in the negative electrode\cite{jiang2020voltage}.
As temperature increases, the OCV hysteresis decreases, especially at low SOC levels. This trend is likely due to faster transport processes and reduced overpotentials\cite{lei2019novel}.

\begin{figure}[H]
    \centering
    \includegraphics[width = 1\textwidth]{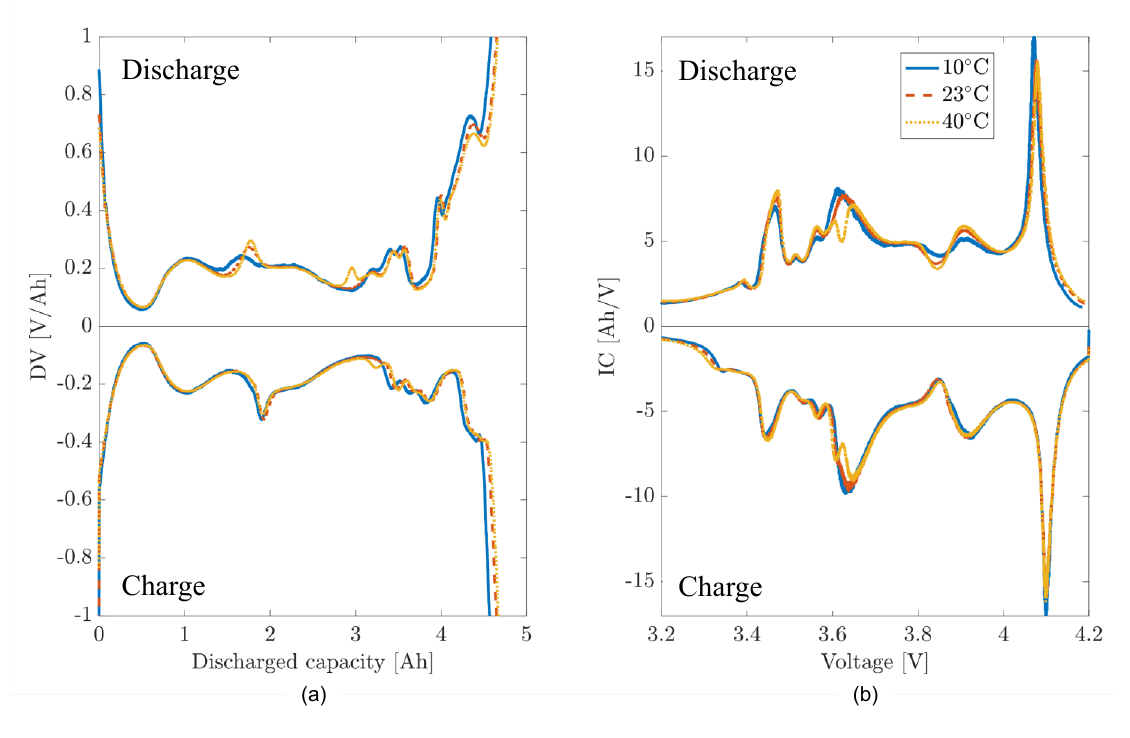}
    \caption{\textbf{Differential voltage (DV) and incremental capacity (IC) curves.} \textbf{(a)} DV and \textbf{(b)} IC curves at 10,  23,  and 40$^\circ$C.}
    \label{fig:dvictemp} 
\end{figure}

Increasing temperature also leads to changes in DV and Incremental Capacity (IC) curves (the IC curve is defined as $1/\mathrm{DV}$).  
Compared to 10$^\circ$C, cells at 23$^\circ$C and 40$^\circ$C have higher charge and discharge capacities\cite{catenaro2021experimental},  which in turn leads to a shift of the DV curves on the right (Supplementary Figure S\ref{fig:dvictemp}(a)).  At 40$^\circ$C,  both DV and IC signatures show one additional peak appearing at \raisebox{2.5pt}{\large\texttildelow}3Ah and \raisebox{2.5pt}{\large\texttildelow}3.6V,  respectively.  While previous studies\cite{hall2018experimental} also note this behavior, the underlying cause remains unclear and warrants further investigation beyond the scope of this paper.

\clearpage

\subsection*{Voltage measurements anomalies in the dataset} \label{sup:IncreasedVoltage}

\begin{figure}[H]
	\centering
	\includegraphics[width=\textwidth]{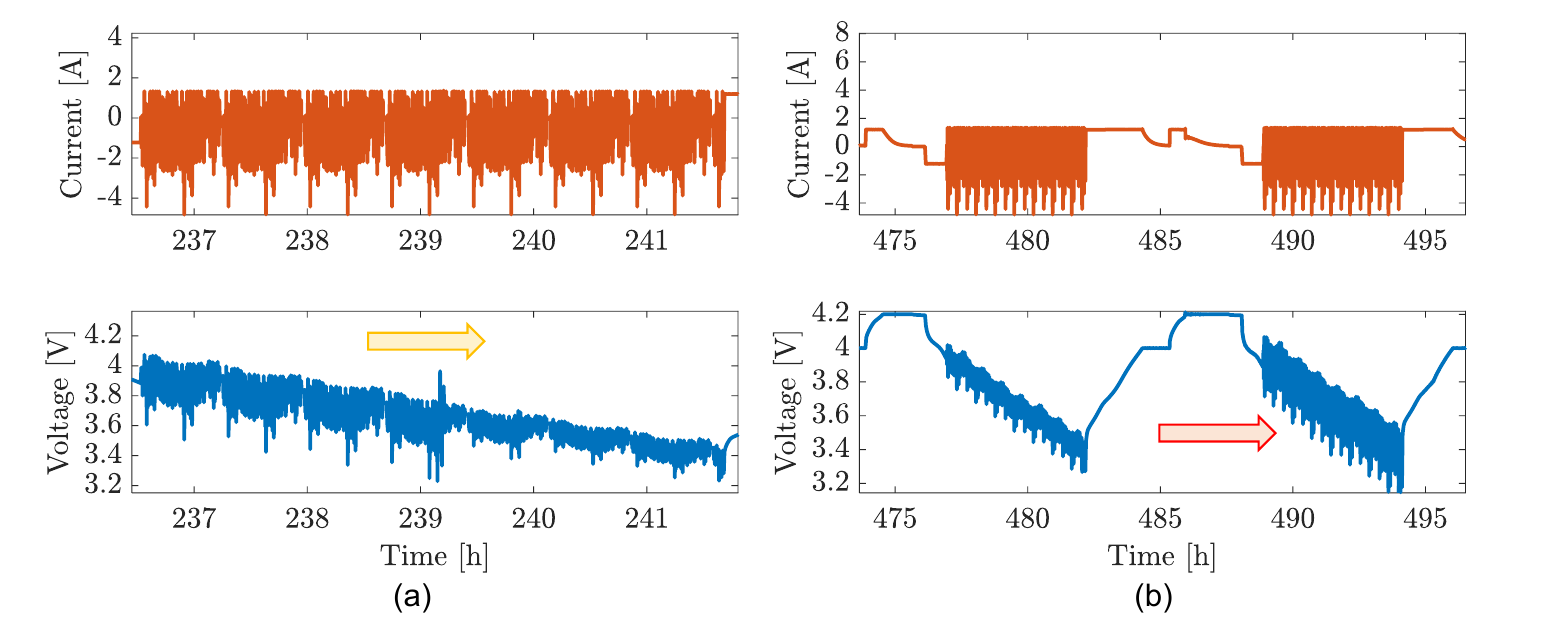}
	\caption{\textbf{cell W7 Voltage Measurments Anomalies.}
	 Example of voltage measurements anomalies for cell W7 within Batch \#2. \textbf{a} Anomalous voltage range change within the same aging cycle. \textbf{b} Wider voltage range in the battery compared to the previous cycle.}
	\label{fig:IncreasedVoltage}
\end{figure}

\begin{figure}[H]
	\centering
	\includegraphics[width=\textwidth]{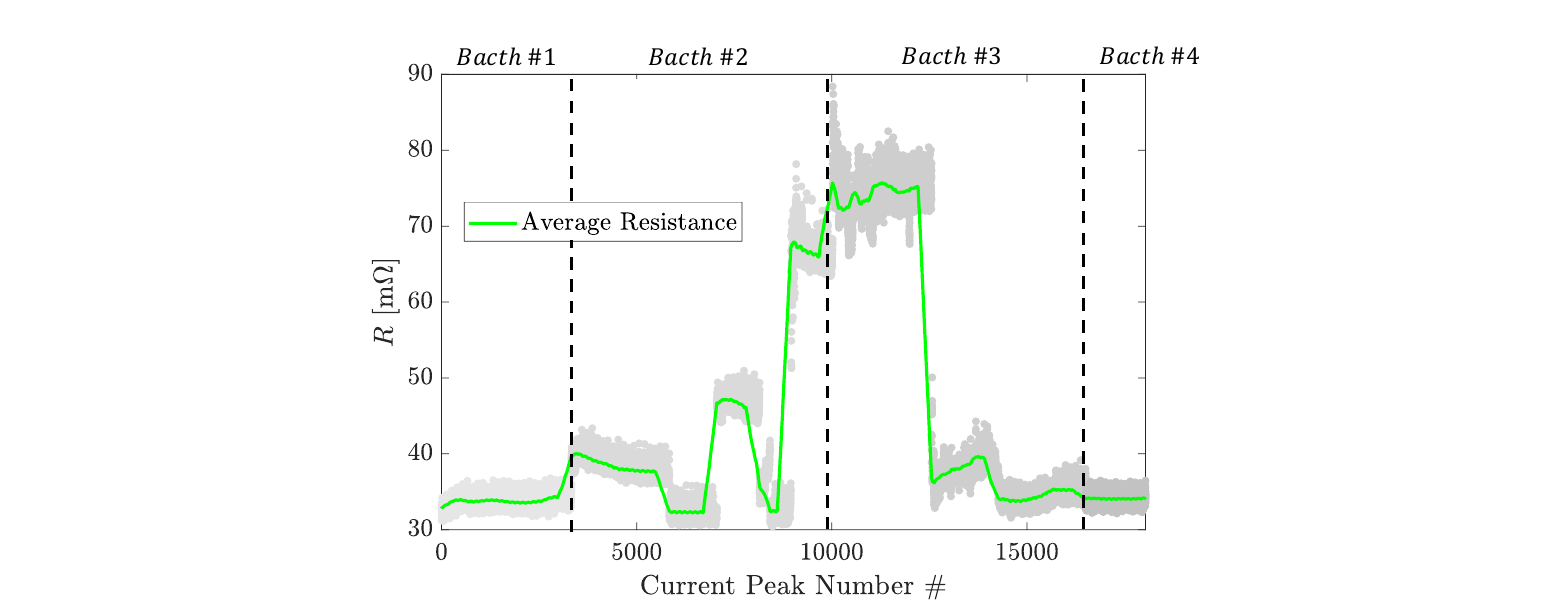}
	\caption{\textbf{Cell W7 Resistance ($R$) during discharge phase over acceleration peaks.}
	 Resistances measured throughout the life of cell W7. In Batches \#2 and \#3 changes in resistance values are observed due to increased voltage.}
	\label{fig:CellW7Resistances}
\end{figure}

As noted in the README file of the dataset paper, cells W4, W5, and W7 exhibited voltage measurement anomalies due to an equipment issue. Supplementary Figure S\ref{fig:IncreasedVoltage}(a) displays an anomalous voltage range change within the same aging cycle, while Supplementary Figure S\ref{fig:IncreasedVoltage}(b) shows a wider voltage range in the battery compared to the previous cycle, despite no changes in the current profile. Additionally, Supplementary Figure S\ref{fig:CellW7Resistances} presents the resistance values for cell W7, which increased in Batches \#2 and \#3 due to the elevated output voltage. For this reason,  the resistance values for cell W7 have been excluded from this study.

\clearpage

\subsection*{Estimation models comparison} \label{sup:model_comparison}

To select the appropriate ML estimation model, we train and test various models to predict SOH and compare their performance. This Supplementary Note evaluates results from different models: linear regression (LRM), feed-forward neural networks (FNN), autoregressive moving average with extra input (ARMAX), and recurrent neural networks (RNN).
While the LRM offers practical benefits, it only captures simple linear relationships between input variables and estimates. FNNs are effective for handling complex non-linear relationships but have drawbacks for real-world applications. They require a large number of training samples to ensure good performance and generalizability. Additionally, FNNs are difficult to interpret due to their black-box nature and numerous parameters.
Since LRM and FNNs are not well-suited for capturing causality, we also tested ARMAX and RNN models. These models can effectively capture causal relationships in battery degradation due to their ability to use past inputs. ARMAX models the output as a linear function of past outputs and current inputs, while RNNs can handle non-linearities but require more computational resources. ARMAX integrates autoregressive (AR) and moving average (MA) models with exogenous inputs (X), defining the relationship between the response and input variables as follows:

\begin{equation}
	A(q)y(t) = B(q)\boldsymbol{u}(t) + C(q)\epsilon(t)
	\label{eq:ARMAX}
\end{equation}

where $q$ is the back-shift operator, $A(q)=1+a_1q^{-1}+...+a_{n_{AR}}q^{-n_{AR}}$ with the parameters $a_i$ of the autoregressive part,
$B(q)=1+b_1q^{-1}+...+b_{n_{X}}q^{-n_{X}}$ with the parameters $b_i$ of the exogenous input part, and 
$C(q)=1+c_1q^{-1}+...+c_{n_{MA}}q^{-n_{MA}}$ with the parameters $c_i$ of the moving average part.
A grid search was carried out to select $n_{AR}$, $n_{X}$, and $n_{MA}$ ARMAX parameters which minimize the root mean squared error (RMSE) obtained over V4 cell data, using W8 cell as the training dataset.
The lowest RMSE was achieved by $n_{AR}=1$,  $n_{X}=0$, and $n_{MA}=2$ between the sets $(0,1,2,3)$, $(0,1,2,3)$, and $(0,1,2,3)$, respectively.
For effective SOH estimation, we optimized hyperparameters in the FNN and RNN training processes, focusing on the number of layers, neurons per layer, and activation functions. A grid search strategy was used, with layer counts from 1 to 3 and neuron counts ranging from 16 to 1024. Activation functions tested included tanh, sigmoid, relu, and identity. For RNNs, we evaluated various recurrent layers available in the Matlab Deep Learning Toolbox, such as LSTM, bidirectional LSTM, gated recurrent, and LSTM projected layers. Additionally, two fully connected layers, each with 64 neurons, were used to enhance the model’s ability to detect non-linearities and recognize patterns.
%Two fully connected layers at the RNN's input and output are used to enhance the model's ability to detect non-linearities and recognize underlying patterns.
%The number of neurons in the first and second fully connected layers are $\textbf{N}=64$, and $\textbf{M}=64$, respectively.
All estimation models were trained using the same dataset from cell W8. The results, tested on cells V4, W7, and W9, are shown in Supplementary Figure S\ref{fig:ResultsEnergies W8}, and RMSEs are quantified in Supplementary Table S\ref{tab:RMSE_W8}. The RMSEs indicate that no single model outperforms the others in terms of accuracy. Thus, the LRM was chosen for its computational efficiency and satisfactory performance.

\begin{figure}[h]
	\centering
	\includegraphics[width = 1\textwidth]{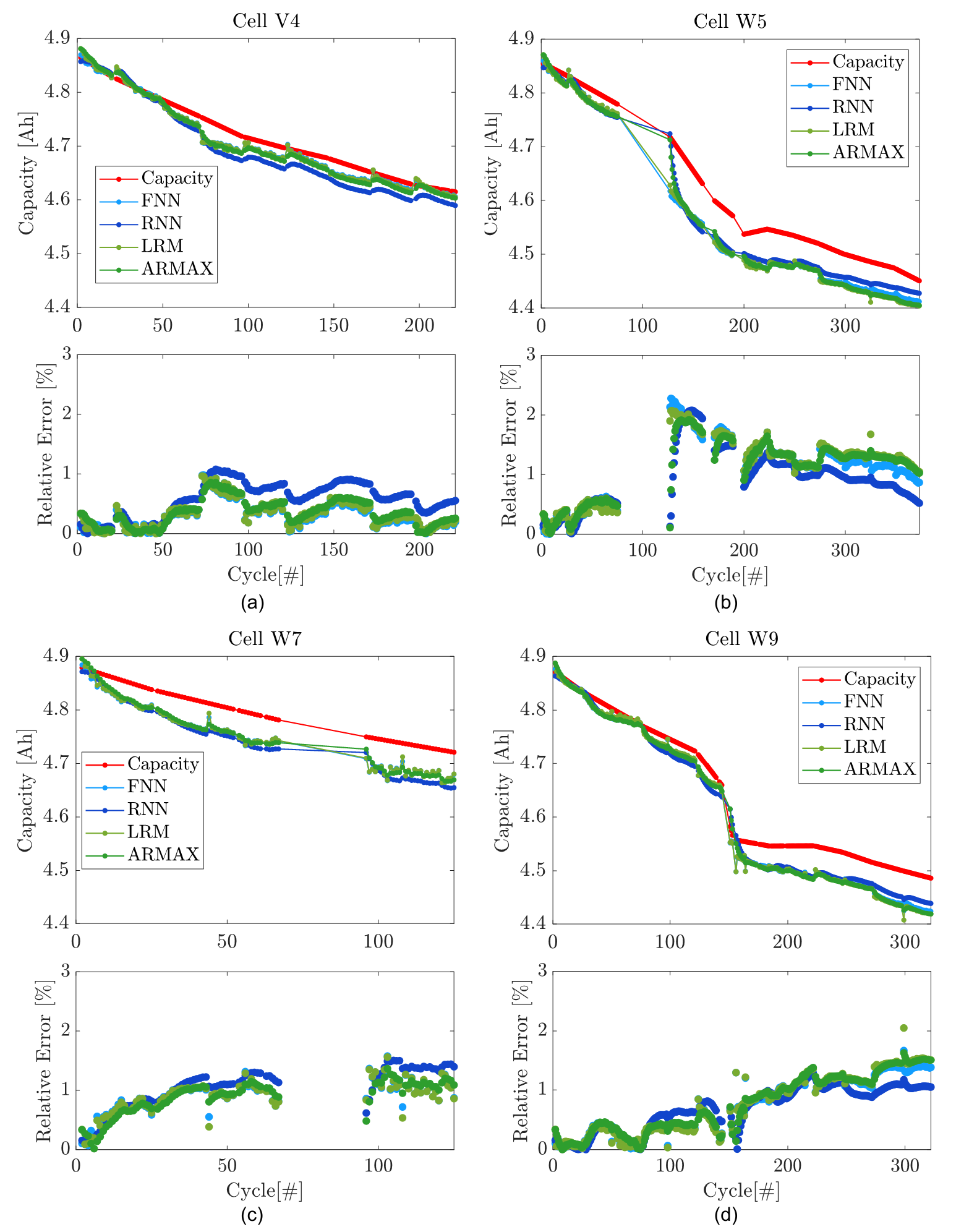}
	\caption{\textbf{SOH estimation results from feed-forward neural networks (FNN), recurrent neural networks (RNN), linear regression (LRM), and autoregressive moving average with extra input (ARMAX) trained on cell W8 using charge and discharge energies as inputs.}
	         The relative error is consistently below 2.5\% for all estimation models. }
	\label{fig:ResultsEnergies W8}
\end{figure}

\begin{table}
	\centering
	\begin{tabular}{lcccc}
		\toprule
		\multirow{2}{*}{\textbf{Estimation model}} & \multicolumn{4}{c}{\textbf{RMSE (\%)}} \\
		\cmidrule(lr){2-5}
		& \it{Cell V4} & \it{Cell W5} & \it{Cell W7} & \it{Cell W9} \\
		\midrule
		FNN   & 0.66 & 1.21 & 0.90 & 0.87 \\
		RNN   & 0.50 & 1.04 & 1.11 & 0.78 \\
		LRM   & 0.40 & 1.27 & 0.94 & 0.91 \\
		ARMAX & 0.39 & 1.23 & 0.93 & 0.91 \\
		\bottomrule
	\end{tabular}
	\vspace{0.5em}
	\caption{\textbf{Estimation error obtained with different models.} Root mean squared errors obtained by FNN, RNN, LRM, and ARMAX when trained on the single W8 cell data and utilizing charge and discharge energies as inputs.}
	\label{tab:RMSE_W8}
\end{table}

\clearpage

\subsection*{Experimental data}\label{sup:experimental_data}
The dataset\cite{pozzato2022lithium} includes aging cycles and RPT data for ten INR21700-M50T cylindrical cells (graphite/silicon anode and NMC cathode) tested at  23$^\circ$C for a period of 30 months.  
A schematic of the cycling phases is shown in Supplementary Figure S\ref{fig:sel_cycle}. Batches of aging cycles are stored in  \texttt{Cycling\_\#} folders;  the number of cycles per folder is listed in Supplementary Table S\ref{suptab1}.
\begin{figure}[h]
	\centering
	\includegraphics[width = 0.9\textwidth]{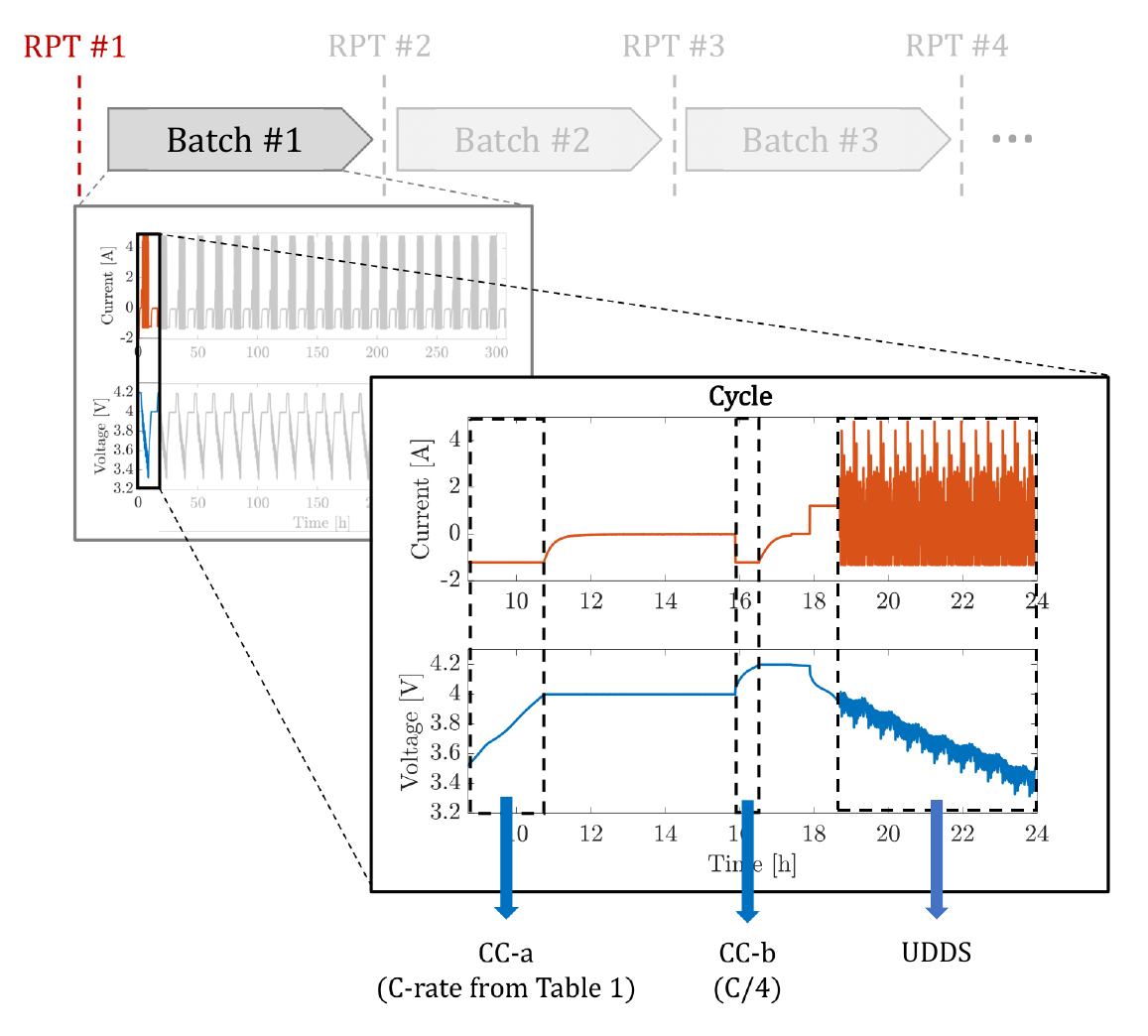}
	\caption{\textbf{Schematic of the experimental campaign.} Aging cycles consist of  charging and a discharging. The standard charging procedure involves four phases: an initial CC-a phase,  CV (at 4 V), a second CC-b phase (at a \textit{C}/4 rate) and final CV at 4.2 V.   Discharging follows the UDDS driving profile. 
	Periodic RPTs assess battery health (RPT \#1,...) and aging cycles performed between the $j^{th}$ and $(j+1)^{th}$ RPTs are grouped into the  $j^{th}$ batch of aging cycles.  Health indicators are extracted from aging cycles data.}
	\label{fig:sel_cycle} 
\end{figure}
\renewcommand{\arraystretch}{1}
\begin{table}[t]
    \centering
    \resizebox{\textwidth}{!}{
    \begin{tabular}{cccccc}
		\multicolumn{6}{c}{\textbf{Aging Cycles}} \\
		\hline
        \textbf{Batch \#} & \textbf{Cell V4} & \textbf{Cell W5} & \textbf{Cell W7} & \textbf{Cell W8} & \textbf{Cell W9} \\ \hline
        \textbf{1} & 20 & 25 & 25 & 25 & 25 \\
        \textbf{2} & 25 & 50 & 50 & 50 & 50 \\
        \textbf{3} & 25 & 50 & 50 & 50 & 47 \\
        \textbf{4} & 25 & 34 & 16 & 23 & 21 \\
        \textbf{5} & 25 & 8 & NaN & 2 & 1 \\
        \textbf{6} & 25 & 20 & NaN & 1 & 1 \\
        \textbf{7} & 25 & 7 & NaN & 6 & 4 \\
        \textbf{8} & 24 & 25 & NaN & 28 & 29 \\
        \textbf{9} & 25 & 25 & NaN & 37 & 37 \\
        \textbf{10} & 25 & 25 & NaN & 25 & 25 \\
        \textbf{11} & NaN & 25 & NaN & 25 & 25 \\
        \textbf{12} & NaN & 25 & NaN & 25 & 25 \\
        \textbf{13} & NaN & 25 & NaN & 25 & 25 \\
        \textbf{14} & NaN & 25 & NaN & 25 & 25 \\
		\hline
        \textbf{Total} & 244 & 369 & 141 & 347 & 340 \\ 
		\hline
    \end{tabular}
    }
    \vspace{0.5em}
    \caption{The table shows for each cell the number of aging cycles within each batch and the total number of aging cycles}
    \label{suptab1}
\end{table}

Data acquisition issues, likely due to sensor malfunctions or tester errors, were observed.  These issues were identified as outliers, with indicator values deviating by more than an order of magnitude from the average. This underscores the utility of these features not only for estimating capacity fade but also for diagnosing data acquisition issues.

\clearpage

\subsection*{Voltage sensitivity for charging impedance computation} \label{sup:voltage_sensitivity_impedance}

The computation of average charging impedances ($Z_{\mathrm{CHG}}$ and $Z_{\mathrm{CHG}_2}$) requires selecting a voltage range $[V_{\mathrm{in}}, V_{\mathrm{fin}}]$. 
Since the choice of voltage range impacts these features, a sensitivity analysis was conducted to identify the voltage range that provides the most useful information.
Three cells (V4, W8 and W9), subject to different \textit{C}-rates during CC-a,  were used as test subjects. The voltage range from 3.6 V to 3.9 V was divided into fifteen sub-intervals, each with a width of 0.05 V and a 0.025 V overlap.

For each cell, the average $Z_{\mathrm{CHG}}^k$ was evaluated across all life cycles, where $k$ represents the $k$-th sub-interval. The correlation between $Z_{\mathrm{CHG}}^k$ and cell capacity was then assessed for each $k$. As shown in Supplementary Figure S\ref{fig:ImpedanceVoltageCorrelation}, the voltage range between 3.8 V and 3.9 V is the most informative, making it the optimal range for $Z_{\mathrm{CHG}}$.

\begin{figure}[H]
	\centering
	\includegraphics[width = 0.7\textwidth]{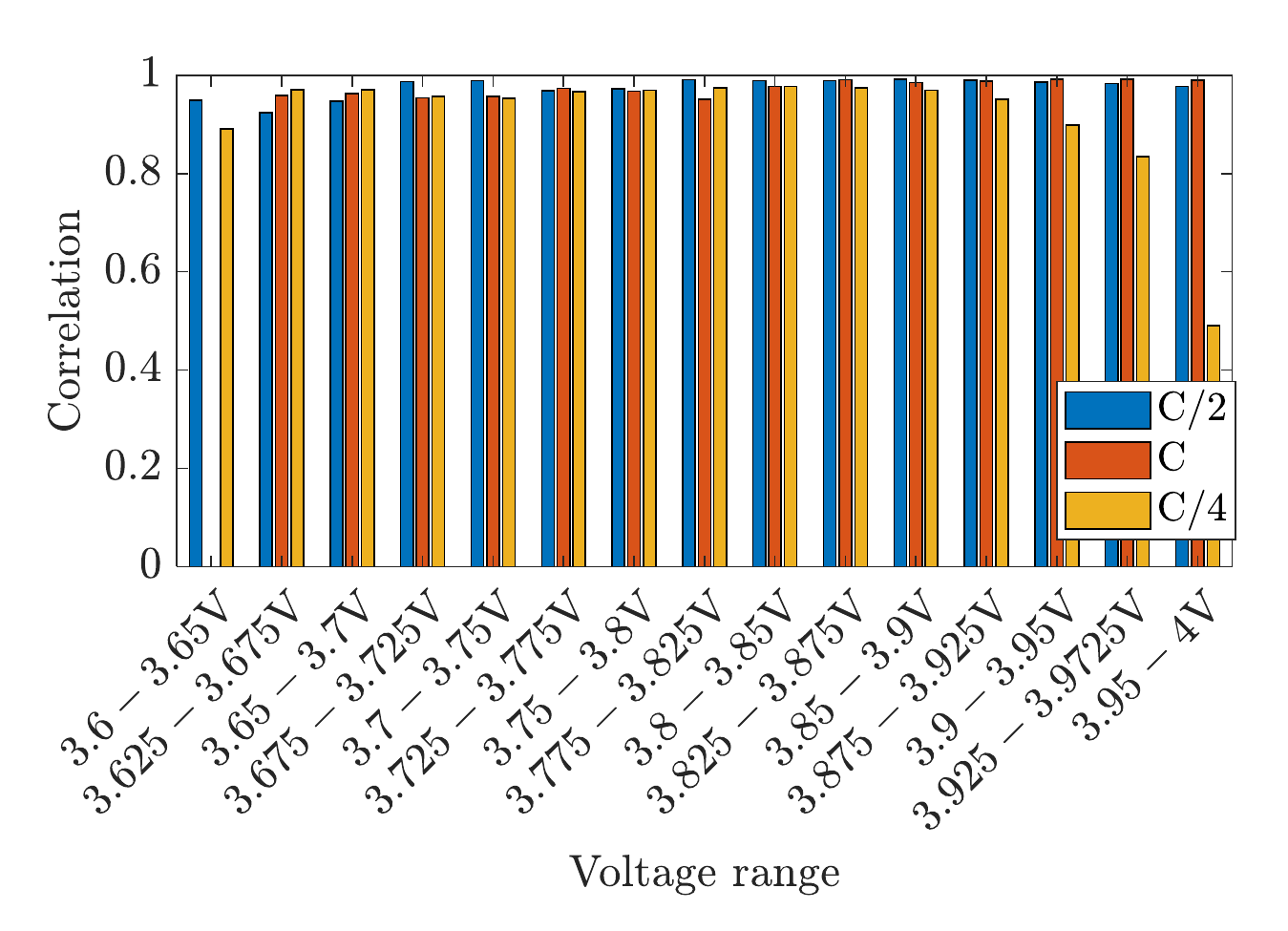}
	\caption{\textbf{Voltage range sensitivity for charging impedance indicator}.  Correlation analysis (i.e., Pearson correlation coefficient)  between average $Z_{\mathrm{CHG}}$ and capacity loss across different voltage sub-intervals for cells charged at $\textit{C}/2$ (blue), $1\textit{C}$ (red) and $\textit{C}/4$ (yellow).}
	\label{fig:ImpedanceVoltageCorrelation} 
\end{figure}

\subsection*{Acceleration peaks extraction from the discharging phase}
\begin{figure}[H]
	\centering
	\includegraphics[width = 1\textwidth]{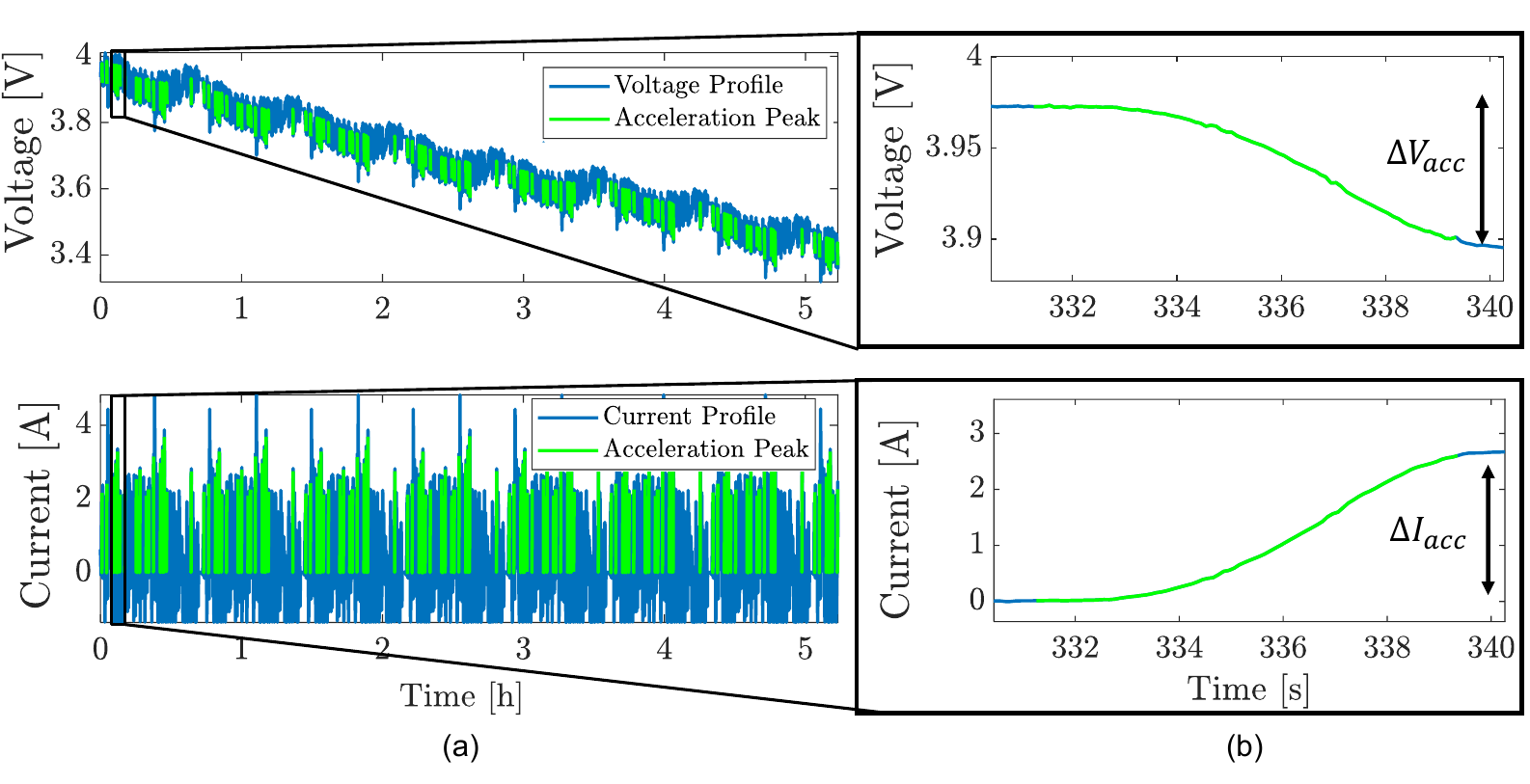}
	\caption{\textbf{Acceleration peaks extraction from the discharging phase.} \textbf{a} Voltage and current profiles during discharging phase (blue). Variations of voltage and current due to acceleration peaks are (green). \textbf{b} Magnified view of a voltage and current variations during acceleration. The battery resistance is evaluated by measuring the ratio between the voltage decrease $\Delta V_{\mathrm{acc}}$ and the current change $\Delta I_{\mathrm{acc}}$ within the same acceleration peak.}
	\label{fig:peakcomputation} 
\end{figure}

\end{document}